\documentclass[10pt,journal]{IEEEtran}
\usepackage{cite}
\usepackage{xcolor}
\usepackage{soul}
\usepackage[utf8]{inputenc}
\usepackage{hyperref}
\usepackage{multirow} 
\usepackage{subfig}
\usepackage{threeparttable}
\usepackage{caption}
\usepackage{amsmath}
\usepackage{amssymb}
\usepackage{pifont}
\usepackage{booktabs}
\usepackage{algorithm}
\usepackage{algorithmic}
\usepackage{lineno,hyperref}
\usepackage{epsfig}
\usepackage{graphicx}
\usepackage{algorithm}
\usepackage{algorithmic}
\usepackage{amsthm}
 \begin{document}

\title{Pseudo-supervised Deep Subspace Clustering}
\author{Juncheng Lv, Zhao Kang, Xiao Lu, Zenglin Xu
\IEEEcompsocitemizethanks{\IEEEcompsocthanksitem J.Lv, Z.Kang, X.Lu are with the School of Computer Science and Engineering, University of Electronic Science and Technology of China, Chengdu, Sichuan, China. email: zkang@uestc.edu.cn}
\IEEEcompsocitemizethanks{\IEEEcompsocthanksitem Z.Xu is with Department of Computer Science and Technology, Harbin Institute of Technology, Shenzhen, China. email: xuzenglin@hit.edu.cn}

}


\IEEEtitleabstractindextext{%

\begin{abstract}
    Auto-Encoder (AE)-based deep subspace clustering (DSC) methods have achieved impressive performance due to the powerful representation extracted using deep neural networks while prioritizing categorical separability. However, self-reconstruction loss of an AE ignores rich useful relation information and might lead to indiscriminative representation, which inevitably degrades the clustering performance. It is also challenging to learn high-level similarity without feeding semantic labels. Another unsolved problem facing DSC is the huge memory cost due to $n\times n$ similarity matrix, which is incurred by the self-expression layer between an encoder and decoder. To tackle these problems, we use pairwise similarity to weigh the reconstruction loss to capture local structure information, while a similarity is learned by the self-expression layer. Pseudo-graphs and pseudo-labels, which allow benefiting from uncertain knowledge acquired during network training, are further employed to supervise similarity learning. Joint learning and iterative training facilitate to obtain an overall optimal solution. Extensive experiments on benchmark datasets demonstrate the superiority of our approach. By combining with the $k$-nearest neighbors algorithm, we further show that our method can address the large-scale and out-of-sample problems. The source code of our method is available: \url{https://github.com/sckangz/SelfsupervisedSC}.
\end{abstract}
}

\maketitle

\IEEEdisplaynontitleabstractindextext

\vspace{1cm}
\IEEEpeerreviewmaketitle

\IEEEraisesectionheading{\section{Introduction}\label{sec:introduction}}

\IEEEPARstart{C}{lustering} is one of the fundamental tasks in computer vision, pattern recognition, and machine learning. Several clustering methods have been developed over the last several decades. Some classical methods such as $k$-means \cite{jain2010data}, agglomerative clustering \cite{johnson1967hierarchical}, and spectral clustering \cite{ng2002spectral} have gained a lot of popularity. In recent years, subspace clustering \cite{agrawal1998automatic} has attracted much attention in handling high-dimensional data. 

In particular, graph-based subspace clustering methods are being investigated by many researchers \cite{elhamifar2013sparse,liu2012robust,tang2020cgd,lv2021multi,peng2016constructing,TangTMM2018,Cao2015Constrained,Yin2018Subspace}. These methods comprise two individual steps. First, the self-expression property, i.e., each data point is expressed as a linear combination of other points, is used to learn an affinity matrix. Then, spectral clustering is applied to the obtained graph matrix. We can see that the performance of subspace clustering depends heavily on the affinity matrix. Thus, different types of regularizer are used for the affinity matrix, such as the nuclear norm \cite{vidal2014low,TangTKDE2019}, $\ell_1$-norm \cite{li2015structured}, $\ell_{2,1}$-norm\cite{Zhenwen2019Learning}, and Frobenius norm \cite{lu2012robust,peng2015robust}. Some recent methods unify graph learning and spectral clustering to benefit from joint learning \cite{kang2021structured,zhan2017graph}. Some methods learn graphs in a latent space to enhance their robustness \cite{patel2015latent,zhang2018generalized}. Kernel subspace clustering is also investigated to explore the nonlinearity of data \cite{yin2016kernel,kang2020structured,xiao2015robust}. Although much progress has been made, shallow methods fall short of reliable discriminative abilities \cite{ma2020towards}.

The tremendous success garnered by deep neural networks (DNNs) has spurred the development of deep clustering. An auto-encoder (AE) is a common building block of many existing deep clustering \cite{xie2016unsupervised,guo2017improved,yang2017towards,zhang2019ae2,kang2020structure}. In essence, an AE harnesses a self-reconstruction loss to find a latent representation, which captures prominent features of raw data. Inappropriate representation might be detrimental to downstream clustering. For example, \cite{huang2014deep}, first, uses an AE to project data into a lower-dimensional space, then apply $k$-means to partition the embedded data. \cite{guo2017improved,xie2016unsupervised,yang2017towards,caron2018deep,jiang2017variational,ghasedi2017deep} perform embedding learning and clustering jointly and alternatively, so that the obtained clustering assignment can supervise the encoding transformation and leads to a clustering-friendly representation. \cite{jabi2019deep,ghasedi2017deep} design some new objective functions by making some strong assumptions 
that are not sufficient for any data set. Improved deep embedding for clustering (IDEC) \cite{guo2017improved} requires a pretraining phase before clustering data and balances the clustering loss with a reconstruction cost. The deep clustering network (DCN) assigns hard labels to samples inducing a discrete optimization process. Deep embedded regularized clustering (DEPICT) \cite{ghasedi2017deep} uses a relative cross-entropy and a regularization term that imposes the size of each cluster relying on some prior knowledge. Unfortunately, the size of clusters is usually unknown for real unsupervised problems. Based on variational AE, variational deep embedding (VaDE) \cite{jiang2017variational} allows coupling clustering with data generation. However, its latent space could be negatively affected since the mean-field approximation causes information loss. Generative adversarial network (GAN)-based deep adversarial clustering \cite{harchaoui2017deep} and information maximizing GANs \cite{chen2016infogan} may conversely affect the true intrinsic structure of data since they change the latent space for clustering. Adversarial deep embedded clustering (ADEC) \cite{mrabah2019adversarial} is proposed to balance the tradeoff between feature randomness and drift; however, it lacks stability because of adversarial training. \cite{shah2017robust} uses a clear continuous objective and integrates cluster number learning and clustering. 

Although AE architecture has achieved far-reaching success for representation learning, it has some drawbacks, along with the sole use of reconstruction loss. For instance, the current AE setting enforces the reconstruction of its input regardless of other data points \cite{wang2017feature,kang2020relation}. In another word, the quality of latent representation could be degraded because of the overlook of rich relation information between neighboring points. Ideally, all points should have different importance in the reconstruction cost to reflect their discriminative roles. Therefore, it is problematic to directly combine clustering and reconstruction since the former aims to destroy non-discriminative details, while the latter loses discriminative information \cite{mrabah2019deep}.


  
Leaning on the recent development of subspace clustering, deep subspace clustering is drawing attention from researchers. Peng et al. \cite{peng2018structured} propose a structured AE for subspace clustering by inputting a prior similarity graph from conventional subspace clustering methods. In essence, their approach belongs to a deep clustering method since it implements $k$-means on latent representation and does not learn any graph. Li et al. \cite{li2017projective} use an encoder to approximate a low-rank graph. A deep subspace clustering (DSC) network \cite{ji2017deep} is the most straightforward implementation of a subspace clustering model in DNN. DSC adds a self-expression layer between the encoder and decoder to learn the similarity graph. Based on DSC, adversarial training is further introduced to boost performance \cite{zhou2018deep}. Recently, multiple fully-connected linear layers have been used to capture low-level and high-level information \cite{kheirandishfard2020multi}. Applications of DSC to multi-view and multimodal data are also considered \cite{abavisani2018deep,zhu2019multi}. Deep clustering is still a challenging task due to the absence of concrete supervision despite the above progress. 

Furthermore, we can observe that the performance of DSC and its variants heavily depends on the quality of latent representations. In addition, self-reconstruction tends to distort relationships between points, which will deteriorate the downstream clustering. Another ignored shortcoming of DSC is that it has a huge memory cost due to the self-expression layer structure, which hinders its applications on large-scale datasets. This explains why all DSC papers use small datasets.

To tackle the above-mentioned issues, in this paper, we develop a pseudo-supervised deep subspace clustering (PSSC) method. The PSSC method aims to learn high-level similarities by feeding pseudo-labels. Specifically, our approach builds upon the general idea of self-training, where a model goes through multiple training iterations. At each iteration, the model uses predictions from the last iteration to relabel unlabeled samples. To achieve this, we introduce a classification module attached to a feature extraction module (encoder) that can use latent representation and similarity graph to construct pseudo-label to supervise the training of feature learning and graph learning \cite{lee2013pseudo}. Furthermore, to preserve the local structure and obtain a more informative representation, we adopt a weighted reconstruction loss instead of the widely used self-reconstruction loss in AE. Specifically, each instance $x_i$ is reconstructed by a set of instances $\hat{x}_j$ weighted by the corresponding relation between them. We adaptively learn them from the self-expression module instead of using fixed relations. 

\begin{figure}[!htp]
\centering
{\includegraphics[width=.48\textwidth]{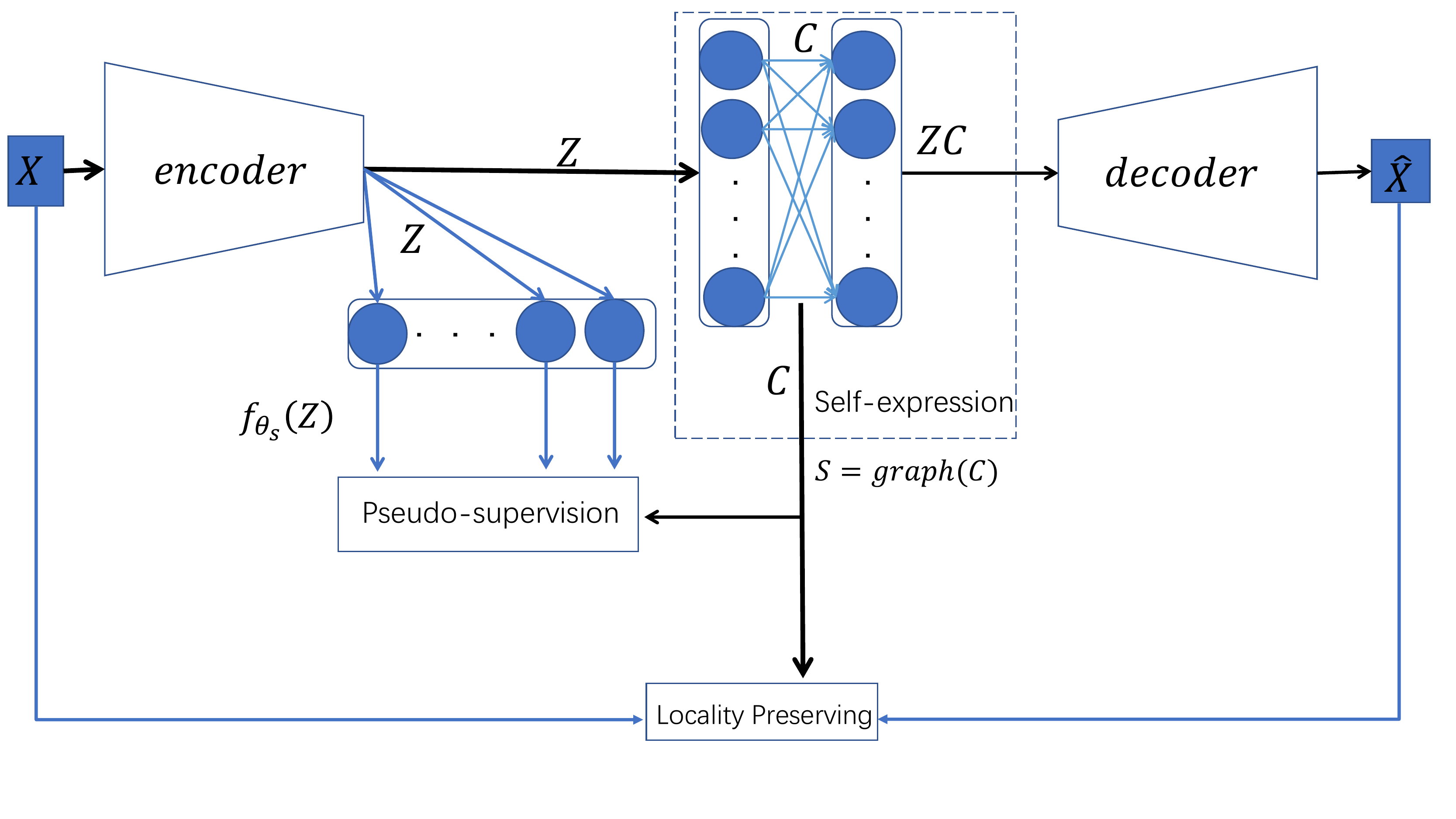}}
\caption{Architecture of pseudo-supervised deep subspace clustering (PSSC). The input $X$ is mapped to $Z$ through an encoder, and $Z$ is self-expressed by $ZC$ and then reconstructed as $\hat{X}$ through a decoder. $f_{\theta_s}(Z)$ is the output of the classifier module. $f_{\theta_s}(Z)$ and $C$ are harnessed to provide pseudo information for self-supervision.}
\label{arch}
\end{figure}

The architecture of our method is shown in Fig.\ref{arch}. The weights of the self-expression layer correspond to the coefﬁcient $C$, i.e., the relations between data samples. Locality preserving ensures the reconstruction of data with less information loss, while the pseudo-supervision guides the training of the entire network.
In summary, our contributions can be summarized as follows:
\begin{itemize}
\item We propose an efﬁcient training framework that integrates DSC and pseudo-supervision. At the same time, it preserves the local neighborhood structure on the data manifold. Feature extraction and similarity learning beneﬁt from such pseudo-supervision since some data with correct labels will propagate useful information, achieving better pseudo-information for supervision. 
\item An approach to address the large-scale data challenge is demonstrated, which is the first to handle 100,000 data points for DSC. 
\item Extensive experiments verify the superiority of our method compared to numerous state-of-the-art subspace models.
\end{itemize}

\section{Related Work}
In this section, we review some works related to the proposed method. 
\subsection{Deep Clustering}
Existing deep clustering methods \cite{xie2016unsupervised,chang2017deep,yang2016joint} mainly aim to combine deep feature learning with traditional clustering methods. AE \cite{bengio2007greedy} is a popular feature learning architecture for unsupervised tasks. It compresses data $X\in\mathcal{R}^{d\times n}$ into a low-dimensional representation $Z\in\mathcal{R}^{m\times n}$ where $m\ll d$, reconstructing the original data. It is often composed of an encoder $F$ and a decoder $G$ with a mirror construction. Their parameters are denoted as $\Theta_e$ and $\Theta_d$, respectively. Specifically, an AE can be optimized by the following objective function:
\begin{equation}
 \min_{\Theta_e,\Theta_d} \sum\limits_{i=1}^{n} \|X_i-G_{\Theta_d}(F_{\Theta_e}(X_i))\|^2=\sum\limits_{i=1}^{n}\|X_i-\hat{X_i}\|^2 .
 \label{autoencoder}
\end{equation}
For the convenience of notation, we let
$\hat{X_i}=G_{\Theta_d}(F_{\Theta_e}(X_i))$. Although this basic model has led to far-reaching success for data representation, it forces the reconstruction of its input without considering other data points in data, resulting in information loss and consequently degrades clustering performance \cite{wang2017feature}. Many methods are proposed to minimize the loss caused by traditional clustering methods to regularize the learning of the latent representation of an AE. In deep manifold clustering \cite{chen2017unsupervised}, the authors interpret the locality of a manifold as similar inputs that should have similar representations and minimize the reconstruction error of $X_i$ itself and its local neighborhood. However, they define reconstruction weights either in a supervised or pre-defined way, and the relations between samples are not flexible for modeling. To get effective representations, other authors manage to keep locality property on the latent space. The deep embedding network \cite{huang2014deep} combines AE and $k$-means while maintaining locality-preserving and group sparsity constraints on latent representations. Two representations are enforced to be similar if their corresponding raw samples are similar, i.e., the graph manifold assumption. The deep embedded clustering (DEC) \cite{xie2016unsupervised} method fine-tunes the encoder by minimizing KL divergence between soft assignment and target distribution. 

To take advantage of the effective self-expression property in traditional subspace clustering, DSC \cite{ji2017deep} implements the subspace clustering method in DNNs by adding a novel self-expressive layer between the encoder and decoder, achieving promising results. More recently, based on the framework of \cite{ji2017deep}, a deep adversarial network with a subspace-speciﬁc generator and a subspace-speciﬁc discriminator are adopted for subspace clustering in \cite{zhou2018deep}. However, the discriminator needs to use the dimension of each subspace, which is usually unknown in practice. By introducing multiple AEs, DSC can be used on multi-view data \cite{zhu2019multi}. Self-paced learning mechanism is also introduced into DSC to boost its generalization ability in \cite{jiang2018learn}. All these methods use the self-reconstruction loss and ignore the local structure information.
\subsection{Pseudo-supervised Learning}
The main challenge for clustering is the lack of labels. Some recent efforts manage to boost clustering performance from two concepts: self-supervision and pseudo-supervision. Self-supervised learning \cite{jing2019self,kolesnikov2019revisiting} generally needs to design a pretext task, where a target objective can be computed without supervision. It assumes that the learned representations of the pretext task contain high-level semantic information that is useful for solving downstream tasks of interest, such as image classiﬁcation. Therefore, the success of self-supervision highly depends on the selected pretext. Some popular pretexts examples include predicting the colorization, predicting unpainted patches, and predicting the spatial relationship of different patches and adversarial objective \cite{kolesnikov2019revisiting,kilinc2018learning}.

Pseudo-supervision trains a model using pseudo-labels rather than ground truth labels \cite{lee2013pseudo}. In contrast to self-supervision, it does not use any proxy. Pseudo-labeling is a more recent self-training algorithm focusing on image classification \cite{oord2018representation}. Pseudo-labeling trains a model using labeled data simultaneously with unlabeled data. The corresponding predictions on unlabeled samples are referred to as pseudo-labels \cite{lee2013pseudo}. Recent research shows that it is possible to conduct clustering with the help of pseudo-labels. Deep Cluster \cite{caron2018deep} uses the pseudo-label computed by $k$-means on output features as supervision to guide the training of DNNs. The deep comprehensive correlation mining (DCCM) \cite{wu2019deep} uses the local robustness assumption and uses pseudo-graphs and pseudo-labels to learn better representations. The self-supervised convolutional subspace clustering \cite{zhang2019self} introduces a spectral clustering module and a classification module into DSC, i.e., using the clustering results to supervise the training of a network. In practice, the main challenge of pseudo-supervision stems from the fact that some self-acquired labels are unreliable and mismatch the real labels. This can mislead data grouping by learning non-representative features, decreasing the discriminative ability of a model. This observation is verified in prior publications \cite{lee2013pseudo,caron2018deep}.
 
 In this paper, we propose a joint optimization framework by combining DSC and pseudo-supervised learning. In this way, useful pseudo-supervision information from latent representation and self-expression can be used to guide feature and similarity learning, refining pseudo-information.

\section{Proposed method}

In this section, we will formally present our proposed PSSC method, a novel pseudo-supervised deep AE network to learn latent representations and sample relations of data. Our network is composed of a locality preserving module, a self-expression module, and a pseudo-supervision module. These three components are seamlessly interconnected and mutually enhance each other.
\subsection{Base Model}

We will introduce the first two modules. In the next subsection, we focus on the pseudo-supervision module.


\textbf{Locality Preserving Module. } To preserve the local structure, we use a weighted reconstruction for an AE instead of Eq.(\ref{autoencoder}). More specifically, $X_i$ is reconstructed by $\hat{X}_j$ with weight $S_{ij}$, where $S_{ij}$ is the similarity between samples $X_i$ and $X_j$. Samples with larger distances between them should have a lower similarity. Then, we obtain the loss function of our weighted reconstruction as follows:
\begin{equation}
L_0={\sum_{ij} S_{ij}||X_i-G_{\Theta_d}(F_{\Theta_e}(X_j))||^2}.
\label{GAE-f}
\end{equation}

Since $S$ is a symmetric matrix, Eq.(\ref{GAE-f}) can be further transformed as

\begin{equation}
\begin{split}
L_0&=\sum S_{ij}||X_i-\hat{X}_j||^2 \\
&=\sum S_{ij}(||X_i||^2-2X_i^T\hat{X}_j+||\hat{X}_j||^2)\\
&=\sum S_{ij}[(||X_i||^2-2X_i^T\hat{X}_i+||\hat{X}_i||^2)\\
\quad&+2(X_i^T\hat{X}_i - X_i^T\hat{X_j})]\\
&=Tr[{(X-\hat{X})}^TD(X-\hat{X})]+ 2Tr(X^TL\hat{X}), 
\end{split}
\end{equation}
where diagonal matrix $D=Diag(\sum_{j=1}^nS_{ij})$, and $L=D-S$ is the Laplacian matrix. We can see that the similarity matrix $S$ would be crucial to the performance of our network. Unlike many existing work using pre-defined values, which often fail to characterize the true relations between samples \cite{kang2020robust}, we propose to automatically learn $S$ from data based on the self-expression concept.

\textbf{Self-Expression Module. }State-of-the-art subspace clustering methods are based on the self-expression property of data, Its basic idea is that each data point can be represented by a linear combination of other data points in the same subspace \cite{elhamifar2013sparse,liu2012robust}. The combination coefficient represents the relation between samples. To learn feature representations that are suitable for subspace clustering, the loss function of  our self-expression module is
\begin{equation}
L_1= \alpha \|C\|_p+\frac{1}{2}\|Z-ZC\|_F^2 \quad s.t.\quad diag(C)=0,
\label{self-expressloss}
\end{equation} 
where the first term is a certain regularization function on $C$, and the second term minimizes the reconstruction error in the latent space $Z$, and $\alpha$ is a tradeoff parameter. Matrix $C$ can represent the subspace structure of data, i.e., $C_{ij}=0$, if the $i$-th sample and $j$-th sample do not lie in the same subspace. The constraint $diag(C)=0$ is optionally used to eliminate a trivial solution of $C=I$. Same as \cite{ji2017deep}, we add a fully-connected layer without a bias between the encoder and decoder, whose weights represent the coefficient matrix $C$, the so-called self-expression layer, as shown in Fig.\ref{arch}.

Once the self-expression coefﬁcient matrix $C$ is obtained, the similarity is usually computed based on $S = \frac{1}{2}(|C| + |C|^T)$. Since the scale of each row and column of the similarity matrix might be different, we use the symmetric normalized Laplacian for scale normalization while maintaining the symmetry of the similarity matrix $S$ \cite{ng2002spectral}. Specifically, the normalized degree matrix $D_n=I$ and normalized Laplacian matrix $L_n=D^{-\frac{1}{2}}LD^{-\frac{1}{2}}$. Then, we have $Tr[{(X-\hat{X})}^TD_n(X-\hat{X})]$ = $\|X-\hat{X}\|_F^2$. So the loss function of our weighted reconstruction is finally transformed as follows
\begin{equation}
    L_0=\|X-\hat{X}\|_F^2+2Tr(X^TL_n\hat{X}).
    \label{localityloss}
\end{equation}
As the second term of $L_0$ has a constraint for $C$, we can remove the first term of $L_1$.

\subsection{Pseudo-Supervision Module}
As previously mentioned, learning clustering-friendly representations without leveraging any supervisory signal is an open and challenging problem. In fact, besides feature information, correlations are also helpful for deep image feature learning \cite{wu2019deep}. 
Inspired by this, we go one step further and fully exploit the above-learned similarity information. Specifically, we introduce pseudo-graph supervision and pseudo-label supervision to guide the network training by constructing pseudo-graphs and pseudo-labels.


\textbf{Pseudo-graph Supervision. }We introduce a classiﬁcation layer parameterized by $\theta_s$ on top of the feature extraction module to produce pseudo-labels. More specifically, we add a fully-connected layer with a softmax function after the encoder, which transforms $Z$ to $f_{\theta_s}(Z)$, where $f_{\theta_s}(z_i)\in \mathbb{R}^{K}$ represents the prediction feature of $z_i$ and $K$ is the number of clusters. It has the following properties: 
\begin{equation}
\begin{split}
    &\sum\limits_{t=1}^K f_{\theta_s}(z_i)_t=1,\forall i=1,...,n;\\
    &f_{\theta_s}(z_i)_t \ge 0,\forall t=1,...,K.   
\end{split}
\end{equation}
In addition, the distribution of $f_{\theta_s}(Z)$ should be consistent with our previously learned similarity graph $S$. Since $S$ is learned in the training process, it is a pseudo-graph by default.   

The loss induced by pseudo-graph supervision can be defined by
\begin{equation}
L_{graph}=\sum\limits_{z_i,z_j\in Z}\mathcal{L}_g(f_{\theta_s(z_i)},f_{\theta_{s(z_j)}};S_{ij}).
   \label{graphloss}
\end{equation}
There are many choices for the loss function $\mathcal{L}_{g}$ in the literature. For example, we can use the contrastive Siamese net loss \cite{bromley1994signature,luo2018smooth} to regularize the distance between two samples and the binary cross-entropy loss \cite{chang2017deep} to evaluate the similarity. It is worth pointing out that our pseudo-graph is different from that in DAC \cite{chang2017deep} and DCCM \cite{wu2019deep}: Unlike previous research that uses a fixed function to calculate pseudo-graphs, we directly learn it from latent representation. This dynamic approach can reveal complex structures hidden in data. In addition, both DAC and DCCM use a threshold, which is hard to select, to find positive pairs. This could involve noisy false positive pairs. By contrast, our $S$ assigns a probability to each pair of points.  

\textbf{Pseudo-label Supervision. }The correlation explored in previous sections is not transitive, i.e., $S_{ij}$ is not deterministic given $S_{ik}$ and $S_{jk}$ and limited to pairwise samples, which could lead to instability in training. Ideally, a pseudo-graph should have $K$-connected components or partitions, which could be regarded as pseudo-labels \cite{kang2021structured,wu2019deep}. This partition would make the optimal solution $\theta_s^*$ in the Eq. (\ref{graphloss}) lead to one-hot prediction, formulating the pseudo-label as
\begin{equation}
    y_i= \arg\max\limits_{k} [f_{\theta_{s}^*}(z_i)]_k,
\end{equation}
where $[.]_k$ represents the $k$-th element of the prediction vector, and its corresponding probability of the predicted pseudo-label is $p_i=\max[f_{\theta_{s}}(z_i)]_k$ .

In practice, it is difficult to achieve optimal $\theta_s^*$ because of non-convexity. Consequently, the prediction $f_{\theta_{s}}(z_i)$ would not follow the one-hot property. To address this problem, we set a large threshold $thres$ for probability $p_i$ to select highly conﬁdent pseudo-labels for supervision: 

\begin{equation}
V_i=
\begin{cases}
1&\text{if $p_i \ge thres$}, \\
0&\text{otherwise}.
\end{cases}
\end{equation}
By doing so, only samples with highly conﬁdent pseudo-labels contribute to network training.
Thus, the pseudo-label supervision loss can be formulated as: 
\begin{equation}
    L_{label}=\sum\limits_{i=1}^n V_i*\mathcal{L}_{l}(f_{\theta_s(z_i)},y_i),
    \label{labelloss}
\end{equation}
where the loss function $\mathcal{L}_{l}$ is often deﬁned by the cross-entropy loss. 
\subsection{The Unified Formulation}
To obtain a unified framework and jointly train the network, we combine loss (\ref{self-expressloss}), (\ref{localityloss}), (\ref{graphloss}), and (\ref{labelloss}) and reach the final objective function of PSSC:


\begin{equation}
\begin{split}
L(\Theta)=&\|X-\hat{X}_{\Theta}\|_F^2+2Tr(X^TL_n\hat{X}_{\Theta})\\
&+\gamma_1||Z_{\Theta_e}-Z_{\Theta_e}C||^2_F+\gamma_2L_{graph}+\gamma_3L_{label}\\
&s.t.\quad diag(C)=0,
\end{split}
\label{objf}
\end{equation}
where $\Theta$ denotes the network parameters, which include encoder parameters $\Theta_e$, self-expression layer parameters $C$, classification layer parameter $\Theta_s$ and decoder parameters $\Theta_d$. Note that, the output $\hat{X}$ of the decoder is a function of {$\Theta_e$, $C$, and $\Theta_d$}. All the unknowns in Eq.(\ref{objf}) are functions of the network parameters. This network can be implemented using neural network frameworks and trained by back-propagation. Once the network architecture is optimized, we obtain the lower-dimensional representation $Z$ and relation matrix $C$.

\begin{table}
\centering
\renewcommand{\arraystretch}{1.1}
\caption{Statistics of the datasets.}
\resizebox{.45\textwidth}{!}{
\begin{tabular}{c|r c c}
\hline
Dataset & Samples & Classes & Dimensions \\
\hline

ORL    &   400& 40 & 32$\times$32  \\

MNIST  &  1,000& 10 & 28$\times$28  \\

Umist  &   480& 20 & 32$\times$32 \\

COIL20 &  1,440& 20 & 32$\times$32  \\

COIL40&  2,880& 40 & 32$\times$32  \\
\hline

\end{tabular}}

\label{dataset}
\end{table}

\section{Subspace Clustering Experiments}
\label{sim-exp}
In this section, we evaluate the effectiveness of the proposed PSSC method.

\begin{table*}[htbp]
\centering
\caption{Network settings for clustering experiments, including the ``kernel size@channels" and size of $C$.}
\renewcommand{\arraystretch}{1.1}
\resizebox{.50\textwidth}{!}{
\begin{tabular}{c|c|c|c|c|c}
\hline
  & ORL & MNIST & Umist & COIL20 & COIL40\\
\hline 

\multirow{3}{4em}{encoder}&5$\times$5@5&5$\times$5@15&5$\times$5@20&3$\times$3@15&3$\times$3@20\\
&3$\times$3@3&3$\times $3@10&3$\times$3@10&-&-\\
&3$\times $3@3&3$\times $3@5&3$\times$3@5&-&-\\
\hline
C &400$\times$400&1000$\times$1000&480$\times$480&1440$\times$1440&2880$\times$2880\\
\hline
\multirow{3}{4em}{decoder}&3$\times$3@3&3$\times$3@5&3$\times$3@5&3$\times$3@15&3$\times$3@20\\
&3$\times $3@3&3$\times $3@10&3$\times$3@10&-&-\\
&5$\times $5@5&5$\times $5@15&5$\times$5@20&-&-\\
\hline
\end{tabular}}

\label{setup}
\end{table*}
\subsection{Datasets}
We conduct experiments on five benchmark datasets that are commonly used in subspace clustering evaluation because of their relatively high dimensionality. There are two face datasets: ORL, Umist; three object datasets: MNIST, COIL20, and COIL100. Following is the detailed information of these datasets.

\begin{itemize}


\item{ORL}: This dataset comprises 40 subjects, and each subject has 10 images taken with varying poses and expressions.

\item{MNIST}: This dataset contains 10 clusters, including handwritten digits 0-9. Each cluster contains 6,000 images for training and 1,000 images for testing, with a size of 28$\times$28 pixels in each image. We use the first 100 images of each digit.

\item{Umist}: This dataset contains 480 images of 20 persons, and each image is taken under very diﬀerent poses. Each image of the dataset is down-sampled to 32$\times$32.

\item{COIL20}: This dataset contains 1,440 gray-scale images of 20 different toys, and the pixels in each image are 32$\times$32.

\item{COIL40}: This dataset is composed of 40 classes of 2,880 data points.

\end{itemize}
The statistics of the datasets are summarized in Table \ref{dataset}.

\begin{table*}[!ht]
\centering
\renewcommand{\arraystretch}{1.3}
\caption{Clustering results on benchmark datasets.}
\resizebox{0.99\textwidth}{!}{
\begin{tabular}{|c|c c c|c c c|c c c|c c c|c c c| }
\hline
Dataset & \multicolumn{3}{c|}{MNIST}  & \multicolumn{3}{c|}{ORL} &  \multicolumn{3}{c|}{COIL20} &  \multicolumn{3}{c|}{COIL40} &\multicolumn{3}{c|}{Umist}  \\
\hline
 Methods$\backslash$Metrices& ACC & NMI & PUR & ACC & NMI & PUR & ACC & NMI & PUR & ACC & NMI & PUR & ACC & NMI & PUR \\
 \hline
SSC    & 0.4530 & 0.4709 & 0.4940 & 0.7425 & 0.8459 & 0.7875 & 0.8631 & 0.8892 & 0.8747 & 0.7191 & 0.8212 & 0.7716 &  0.6904 & 0.7489 & 0.6554\\
\hline
ENSC   & 0.4983 & 0.5495 & 0.5483 & 0.7525 & 0.8540 & 0.7950 & 0.8760 & 0.8952 & 0.8892 & 0.7426 & 0.8380 & 0.7924 & 0.6931 & 0.7569 & 0.6628\\ 
\hline

KSSC   & 0.5220 & 0.5623 & 0.5810 & 0.7143 & 0.8070 & 0.7513 & 0.7087 & 0.8243 & 0.7497 & 0.6549 & 0.7888 & 0.7284 & 0.6531 & 0.7377 & 0.6256 \\
\hline

SSC-OMP& 0.3400 & 0.3272 & 0.3560 & 0.7100 & 0.7952 & 0.7463 & 0.6410 & 0.7412 & 0.6667 & 0.4431 & 0.6545 & 0.5250 & 0.6438 & 0.7068 & 0.6171\\
\hline

EDSC   & 0.5650 & 0.5752 & 0.6120 &  0.7038 & 0.7799 & 0.7138 & 0.8371 & 0.8828 & 0.8585 & 0.6870 & 0.8139 & 0.7469 & 0.6937 & 0.7522 & 0.6683 \\
\hline

LRR    & 0.5386 & 0.5632 & 0.5684 & 0.8100 & 0.8603 & 0.8225 & 0.8118 & 0.8747 & 0.8361 & 0.6493 & 0.7828 & 0.7109 & 0.6979 & 0.7630 & 0.6670 \\
\hline

LRSC   & 0.5140 & 0.5576 & 0.5550 &  0.7200 & 0.8156 & 0.7542 & 0.7416 & 0.8452 & 0.7937 & 0.6327 & 0.7737 & 0.6981 & 0.6729 & 0.7498 & 0.6562\\
\hline

AE+SSC & 0.4840 & 0.5337 & 0.5290 &  0.7563 & 0.8555 & 0.7950 & 0.8711 & 0.8990 & 0.8901 & 0.7391 & 0.8318 & 0.7840 & 0.7042 & 0.7515 & 0.6785 \\
\hline

DSC-L1 & 0.7280 & 0.7217 & 0.7890 &  0.8550 & 0.9023 & 0.8585 & 0.9314 & 0.9353 & 0.9306 & 0.8003 & 0.8852 & 0.8646 & 0.7242 & 0.7556 & 0.7204\\
\hline

DSC-L2 & 0.7500 & 0.7319 & 0.7991 &  0.8600 & 0.9034 & 0.8625 & 0.9368 & 0.9408 & 0.9397 & 0.8075 & 0.8941 & 0.8740 & 0.7312 & 0.7662& 0.7276\\
\hline

DEC    & 0.6120 & 0.5743 & 0.6320 & 0.5175  & 0.7449 & 0.5400 & 0.7215 & 0.8007  & 0.6931 & 0.4872 & 0.7417 & 0.4163 & 0.5521 & 0.7125 & 0.5917\\
\hline

DKM    & 0.5332 & 0.5002 & 0.5647 & 0.4682 & 0.7332 & 0.4752 & 0.6651 & 0.7971 & 0.6964 & 0.5812 & 0.7840 & 0.6367 & 0.5106 & 0.7249 & 0.5685 \\
\hline

DCCM   & 0.4020 & 0.3468 & 0.4370 &  0.6250 & 0.7906 & 0.5975 & 0.8021 & 0.8639 & 0.7889 & 0.7691 & 0.8890 & 0.7663 & 0.5458 & 0.7440 & 0.5854\\
\hline

DEPICT & 0.4240 & 0.4236 & 0.3560 & 0.2800 & 0.5764 & 0.1450 & 0.8618 & 0.9266 & 0.8319 & 0.8073 & \textbf{0.9291} & 0.8191 & 0.4521 & 0.6329 & 0.4167\\
\hline

DSCDAN & 0.7450 & 0.7110 & 0.7480 & 0.7950 & 0.9135 & 0.8025 & 0.7868 & 0.9131 & 0.7819 & 0.7385 & 0.8940 & 0.7726 & 0.6937 & \textbf{0.8816} & 0.7167\\
\hline

PSSC$_l$& 0.7850 & 0.7276 & 0.7860 & 0.8525 & 0.9258 & 0.8775 & 0.9583 & 0.9667 & 0.9853 & 0.8153 & 0.9240 & 0.8538 & 0.7271 & 0.8525 & 0.7750\\ 
\hline

PSSC& \textbf{0.8430} &\textbf{0.7676} & \textbf{0.8430} & \textbf{0.8675} & \textbf{0.9349} & \textbf{0.8925} & \textbf{0.9722} & \textbf{0.9779} & \textbf{0.9722} & \textbf{0.8358} &   0.9258        & \textbf{0.8642} & \textbf{0.7917} &   0.8670        & \textbf{0.8146}\\
\hline

\end{tabular}}

\label{cluster_result}
\end{table*}

\subsection{Comparison Methods}
We compare our methods with both shallow and deep clustering techniques, including low-rank representation (LRR) \cite{liu2013robust}, low-rank subspace clustering (LRSC) \cite{vidal2014low}, sparse subspace clustering (SSC) \cite{elhamifar2013sparse}, kernel SSC (KSSC) \cite{patel2014kernel}, SSC by orthogonal matching pursuit (SSC-OMP) \cite{you2016scalable}, efficient dense subspace clustering (EDSC) \cite{ji2014efficient}, SSC with pre-trained convolutional AE features (AE+SSC), DSC network with $\ell1$-norm (DSC-L1) \cite{ji2017deep}, DSC network with $\ell2$-norm (DSC-L2), deep embedding clustering (DEC) \cite{xie2016unsupervised}, deep $k$-means (DKM) \cite{fard2018deep}, deep comprehensive correlation mining (DCCM) \cite{wu2019deep}, deep embedded regularized clustering (DEPICT) \cite{ghasedi2017deep}, and deep spectral clustering using dual AE network (DSCDAN) \cite{yang2019deep}.

To show the impact of pseudo-supervision, we make an ablation study by removing $L_{graph}$ and $L_{label}$ in Eq. (\ref{objf}) and term the left model as PSSC$_l$.

\subsection{ Setup}
In our experiments, we use a one-layer convolutional network for the encoder and decoder in COIL20 and COIL40, a three-layer convolutional network for others. The architecture details of the networks are shown in Table \ref{setup}. We choose ReLU as the non-linear activation function and implement our method with TensorFlow. 

First, we pre-train the encoder and decoder without the self-expression layer and softmax layer. Then, we fine-tune the entire network and grid searching for $\gamma_1,\gamma_2,\gamma_3$. The learning rate is set as 1e-3 in pre-training and 1e-4 in the fine-tuning stage. The contrastive Siamese net loss and cross-entropy loss are adopted for $L_{graph}$ and $L_{label}$ respectively. For pseudo-labels, threshold $thres$ is empirically set to 0.8. We use Adam optimizer \cite{kingma2014adam} as the optimizer. The training process is summarized in Algorithm \ref{alg:train}.

\begin{algorithm}
\caption{PSSC} 
\label{alg:affinity} 
\begin{algorithmic}[1]

\REQUIRE
The data set, $X$;\\
the number of clusters, $k$;\\
\ENSURE
The clustering result $Y$;  
\STATE Random initialize the auto-encoder with $\Theta_e^0$, $\Theta_d^0$;
\STATE Pre-train the auto-encoder with all data to obtain parameter $\Theta_e^1$, $\Theta_d^1$;
\STATE Initialize the auto-encoder part of PSSC network with $\Theta_e^1$, $\Theta_d^1$;
\STATE Random initialize self-expression layer parameter $C$;
\STATE \textbf{While} not converge or reach the maximum training epoches \textbf{do} ;
\STATE Train the whole PSSC network;
\STATE Calculate the gradient of the parameters and update the $\quad$parameters through the Adam optimizer;
\STATE \textbf{End while};
\STATE Use $C$ to calculate the affinity matrix $A$ through algorithm \ref{alg:affinity};
\STATE Conduct spectral clustering on $A$ to obtain the clustering result $Y$.

\label{alg:train}
\end{algorithmic}
\end{algorithm}


Similar to DSC \cite{ji2017deep}, after we obtain the relation $C$, we use it to construct an afﬁnity matrix for spectral clustering \cite{ng2002spectral}. To enhance the block-structure and improve clustering accuracy, a typical method employed by various subspace clustering methods is summarized in Algorithm \ref{alg:affinity}, where $\alpha$ is empirically selected according to the level of noise and $q$ is the maximal intrinsic dimension of subspaces \cite{ji2014efficient}. Specifically, $q$ can be determined by the number of nonzero singular values of data points \cite{elhamifar2013sparse}. For these parameters, we use the same setting as our main competitor DSC \cite{ji2017deep}. To quantitatively assess clustering performance, we adopt three popular evaluation metrics: accuracy (ACC), normalized mutual information (NMI), and purity (PUR). 
\begin{algorithm}
\caption{Compute affinity matrix.} 
\label{alg:affinity} 
\begin{algorithmic}[1]
\REQUIRE
The relation matrix, $C$;\\
The number of clusters, $k$;\\
The intrinsic dimension of subspaces, $q$;
\ENSURE
The affinity matrix, $A$;  
\STATE Let $S = \frac{1}{2}(|C| + |C|^T)$;
\STATE Compute the SVD of $S$, $S = U\Sigma V^T$;
\STATE Let $Z = U_m\Sigma_m^{\frac{1}{2}}$, where $m=k*q+1$;
\STATE Compute affinity matrix $A = [ZZ^T]^\alpha$;
\end{algorithmic}
\end{algorithm}

\subsection{Results}
\begin{figure}[!ht]
\centering
\subfloat[PSSC\_MNIST\label{A-MNIST}]{\includegraphics[width=.23\textwidth]{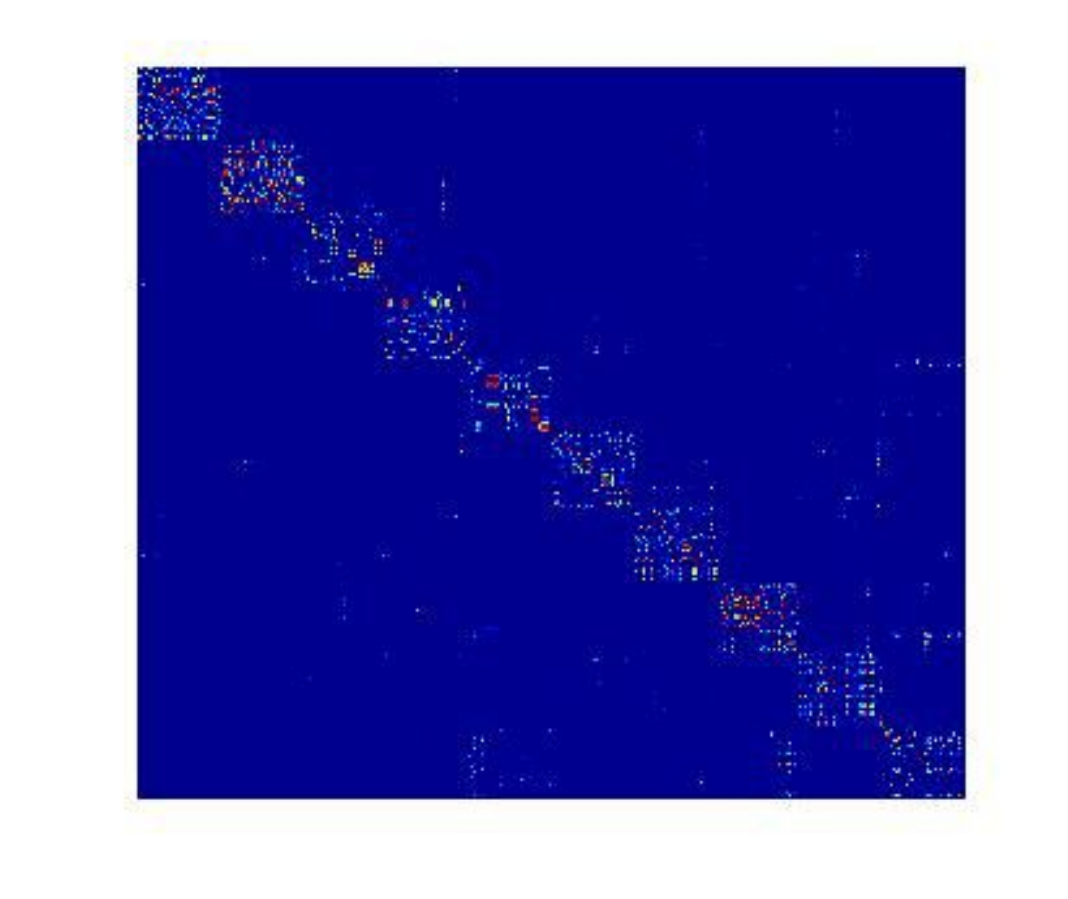}}
\hspace{.05cm}
\subfloat[PSSC\_ORL\label{A-L2}]{\includegraphics[width=.23\textwidth]{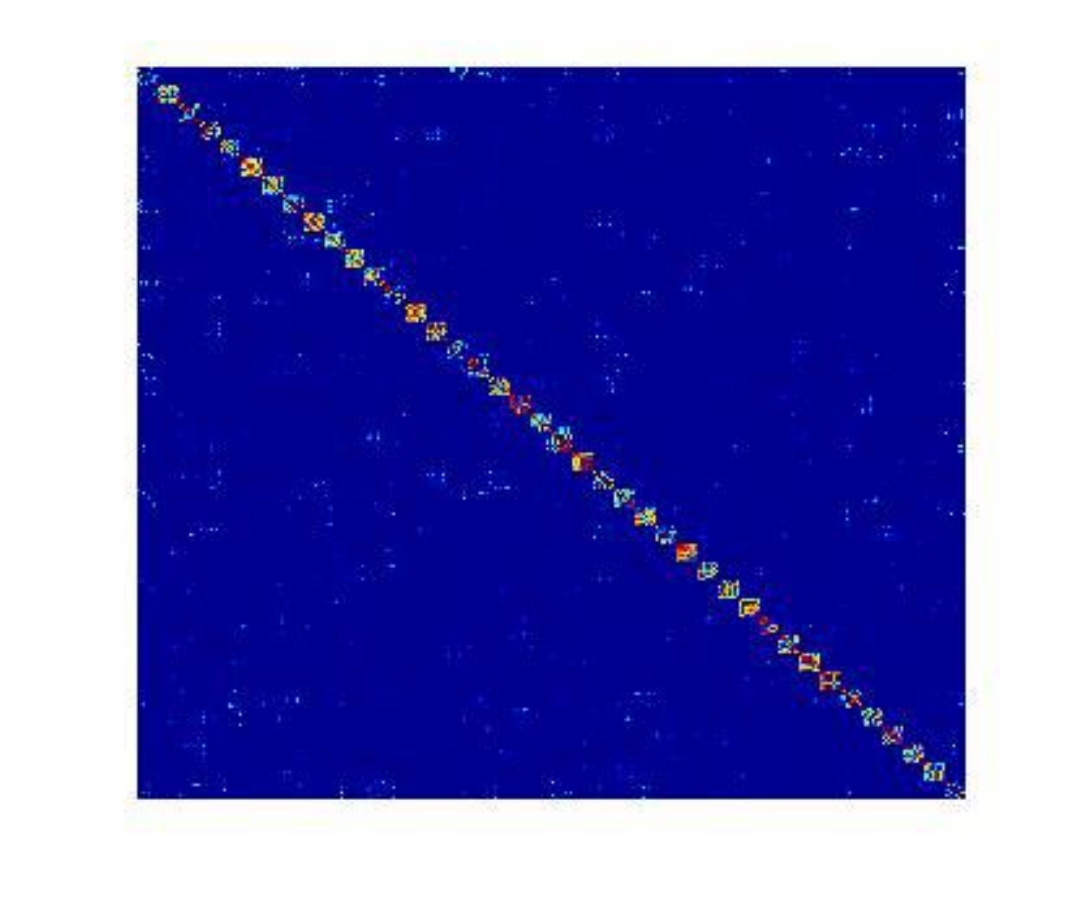}}\\
\subfloat[DSC\_MNIST\label{A-MNIST}]{\includegraphics[width=.23\textwidth]{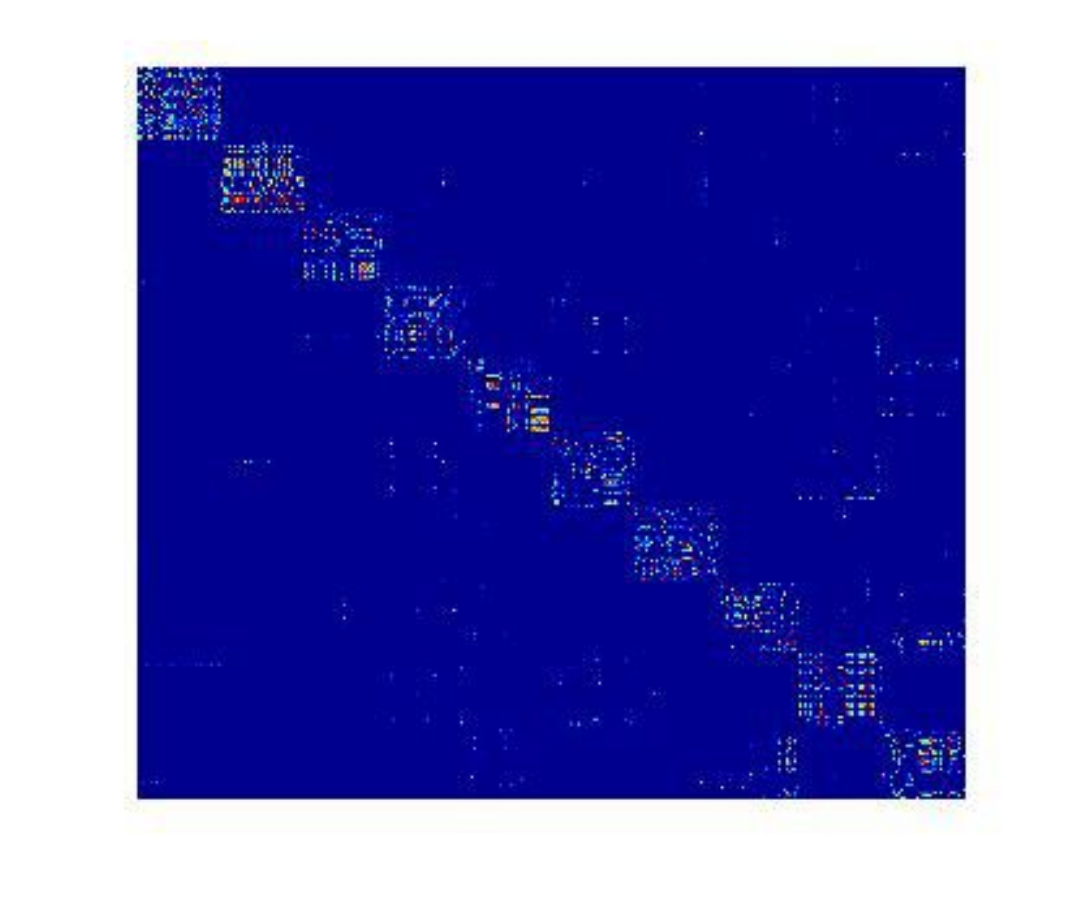}}
\hspace{.05cm}
\subfloat[DSC\_ORL\label{A-L2}]{\includegraphics[width=.23\textwidth]{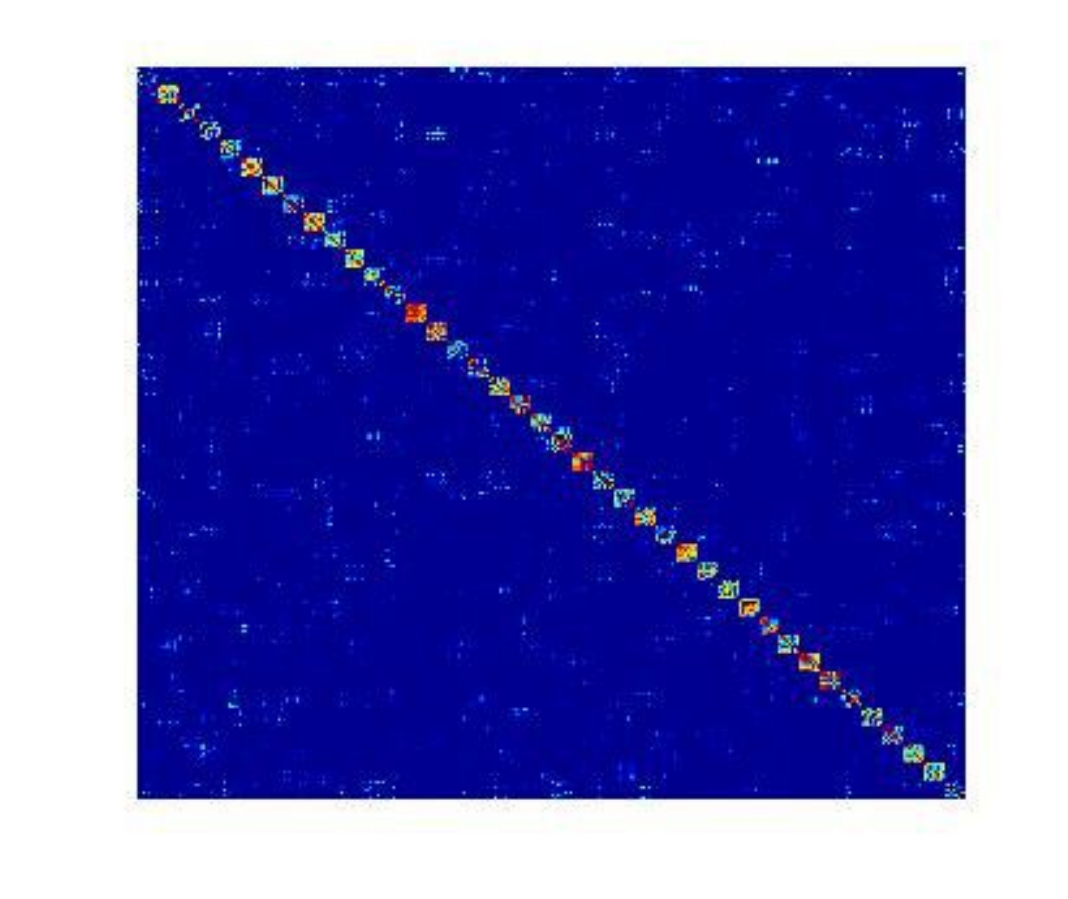}}

\caption{Visualization of learned affinity matrix $A$ on MNIST and ORL dataset. Block-diagonal structures can be observed.}
\label{SimilarityA}
\end{figure}

Table \ref{cluster_result} shows the clustering results on the benchmark datasets. For each measure, the best methods are highlighted in boldface. We can observe that our method achieves the best performance among all comparison methods. In some cases, PSSC outperforms others by a signiﬁcant margin. Specifically, we have the following observations.
\begin{itemize}
\item PSSC outperforms shallow subspace clustering methods considerably. This is consistent with our expectation and mainly attributes to the powerful representation ability of neural networks. 
\item PSSC and PSSC$_l$ outperform the DSC and other deep clustering methods. In particular, the improvement of PSSC$_l$ over DSC veriﬁes the importance of locality preserving since the only difference between these two models stems from AE loss. PSSC enhances DSC by 4.48\%, 4.74\%, and 3.73\% on average in terms of ACC, NMI, and purity, respectively.
\item Compared with PSSC$_l$, PSSC always achieves better results, indicating that pseudo-supervision guides feature and relation learning. For instance, with respect to PSSC$_l$, supervision boosts ACC by 5.8\% and 6.46\% on MNIST and Umist. This demonstrates that pseudo-graphs and pseudo-labels are promising methods for unsupervised tasks.
\item In some cases, DEC, DKM, and DCCM perform even worse than shallow approaches. This is because that they use Euclidean distance or cosine distance to evaluate the pairwise relation, which fails to capture the complex manifold structure. In general, the subspace learning approach works much better in this situation because it is originally designed to handle high-dimensional data. In general, the subspace learning approach works much better in this situation. 
\item Compared to our method, DEPICT \cite{ghasedi2017deep} and DSCDAN \cite{yang2019deep} produce inferior performance. With respect to DEPICT, the performance of DSCDAN is more stable.
\end{itemize}


Furthermore, self-supervised convolutional subspace clustering network ($S^2$ConvSCN-$\ell_2$) \cite{zhang2019self} and deep subspace clustering (DeepSC) \cite{peng2020deep} are two recent DSC methods. Since their authors have not published the source code, we directly cite their results on the datasets we used. For example, on COIL20, our proposed method's ACC (NMI) is 0.9722 (0.9779), while DeepSC achieves 0.9801 (0.9791) and $S^2$ConvSCN-$\ell_2$ reaches 0.9767. We can see that the performance of these methods is close. On ORL, our proposed method's ACC is 0.8675, while $S^2$ConvSCN-$\ell_2$ achieves 0.8875. Hence, our method is comparable to $S^2$ConvSCN-$\ell_2$ and DeepSC. In fact, both $S^2$ConvSCN-$\ell_2$ and DeepSC are not suitable for practical applications since the former involves spectral clustering and the latter needs to solve $\ell_1$ problem in each epoch, which results in high time complexity. By contrast, our pseudo-supervision is simple yet effective.

Furthermore, to intuitively show the similarity learning merit of our approach, we visualize the afﬁnity matrix $A$ of PSSC and DSC in Fig.\ref{SimilarityA},
where $A_{ij}$ indicates the similarity between $Z_i$ and $Z_j$ and brighter pixel means higher similarity. As observed, most of the energy is perfectly located at the block-diagonal pixels. Compared to PSSC, more points of DSC appear in non-diagonal area, which means that they are inaccurately calculated or noisy. This partially explain the good performance of our method.

\begin{figure}[!htp]
\centering
{\includegraphics[width=.48\textwidth]{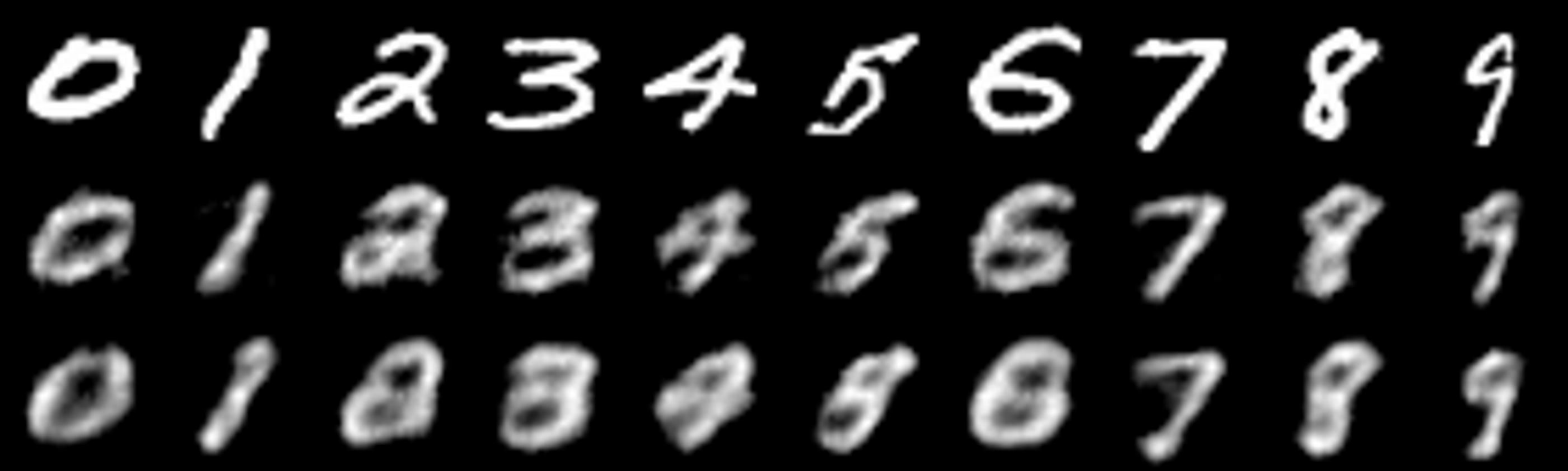}}
\caption{The reconstructed images of MNIST. From top to bottom: original image, PSSC, DSC. }
\label{reconstructed}
\end{figure}

Fig.\ref{reconstructed} shows some reconstructed images on MNIST by our method and DSC. The first row is the original images, the second row is the images reconstructed by our PSSC, and third is DSC. We can find out that PSSC recovers the edges pretty well, while the images of DSC are blurred. This is also validated by PSNR value, PSSC achieves 15.54 while DSC gives 15.04. This verifies the drawback of existing self-reconstruction loss commonly used in AE.

\begin{figure*}[!htbp]
\centering
\subfloat[$\gamma_1$=1]{\includegraphics[width=.31\textwidth]{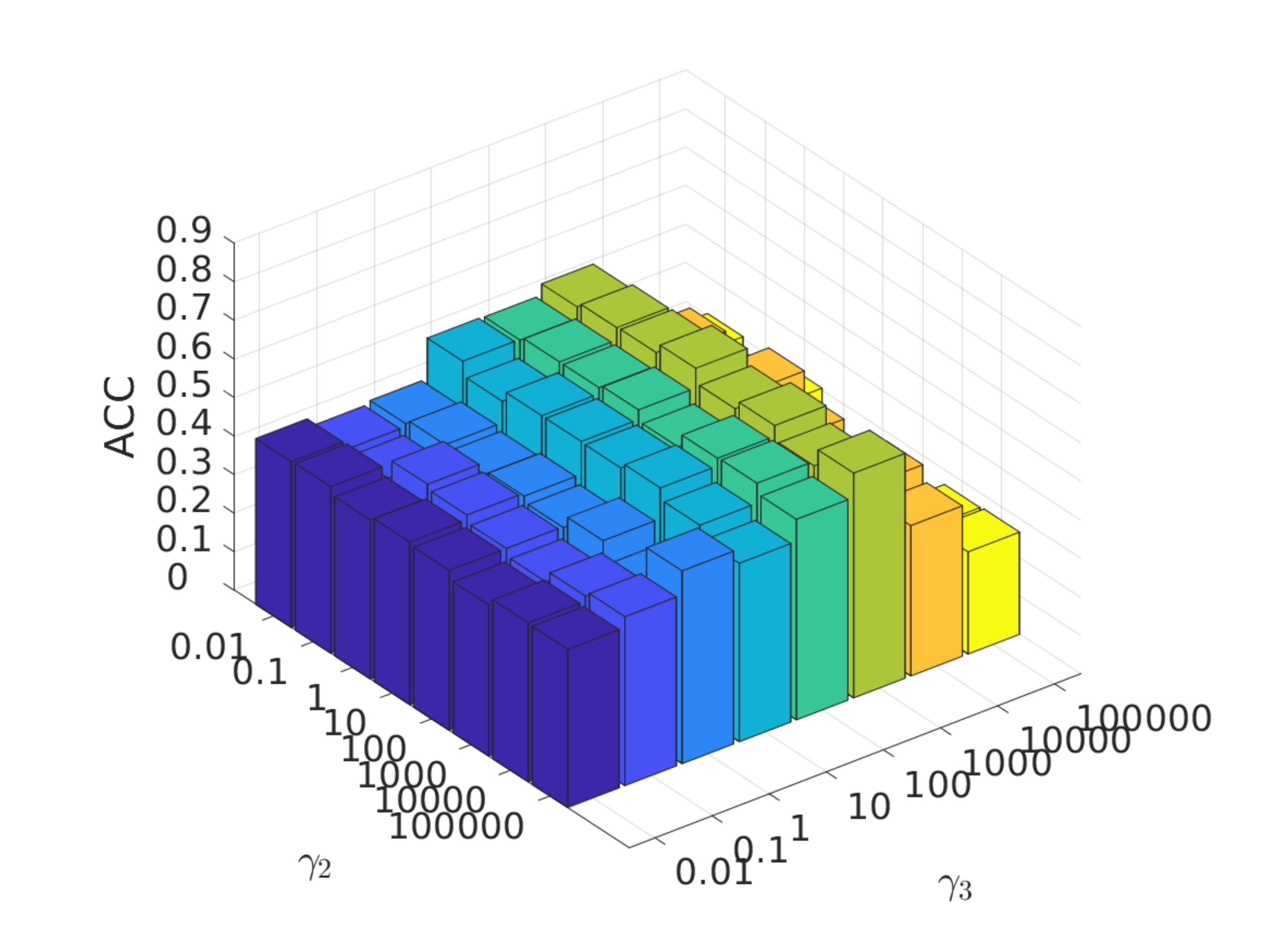}}
\subfloat[$\gamma_1$=1e2]{\includegraphics[width=.31\textwidth]{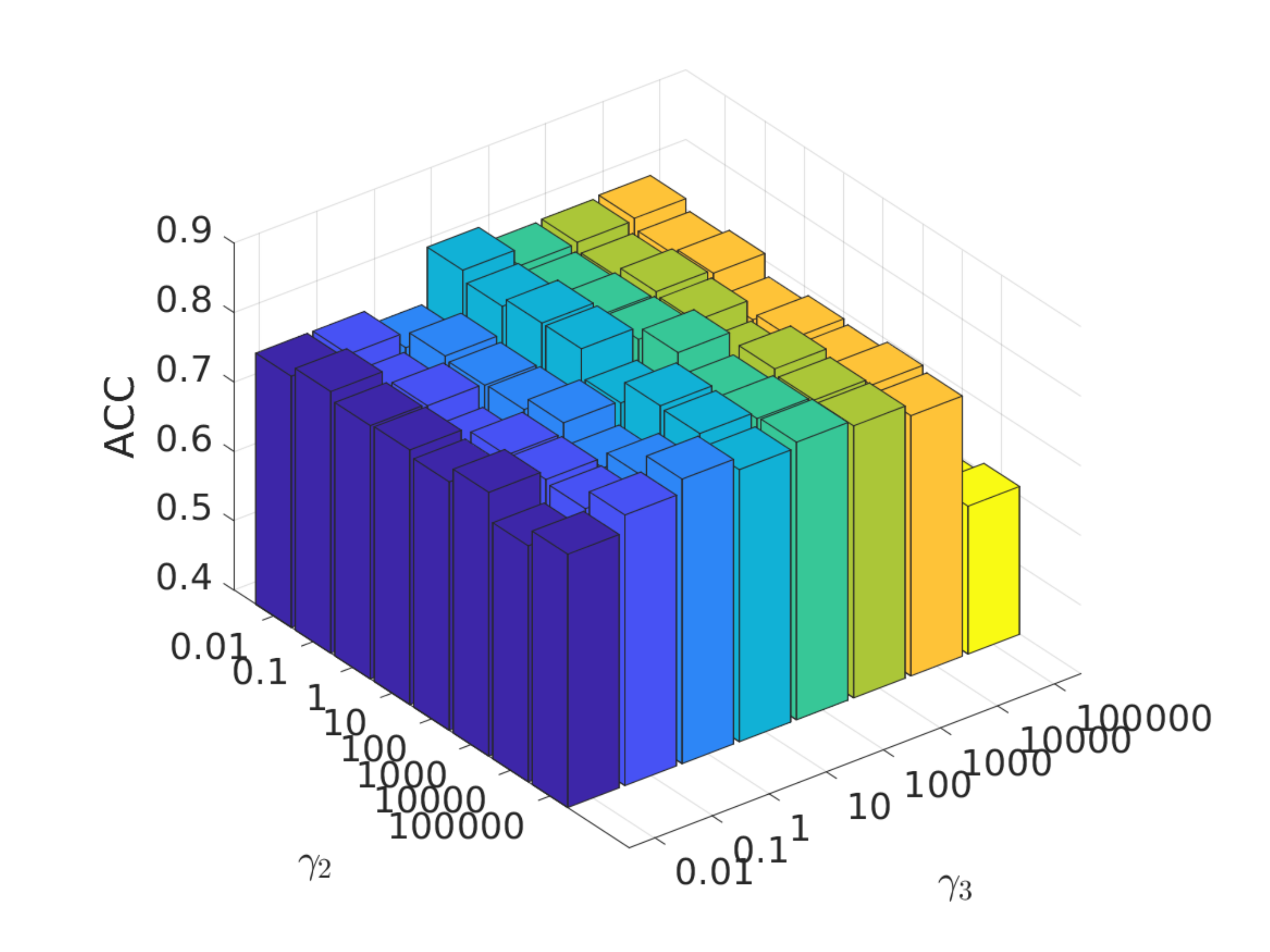}}
\subfloat[$\gamma_1$=1e4]{\includegraphics[width=.31\textwidth]{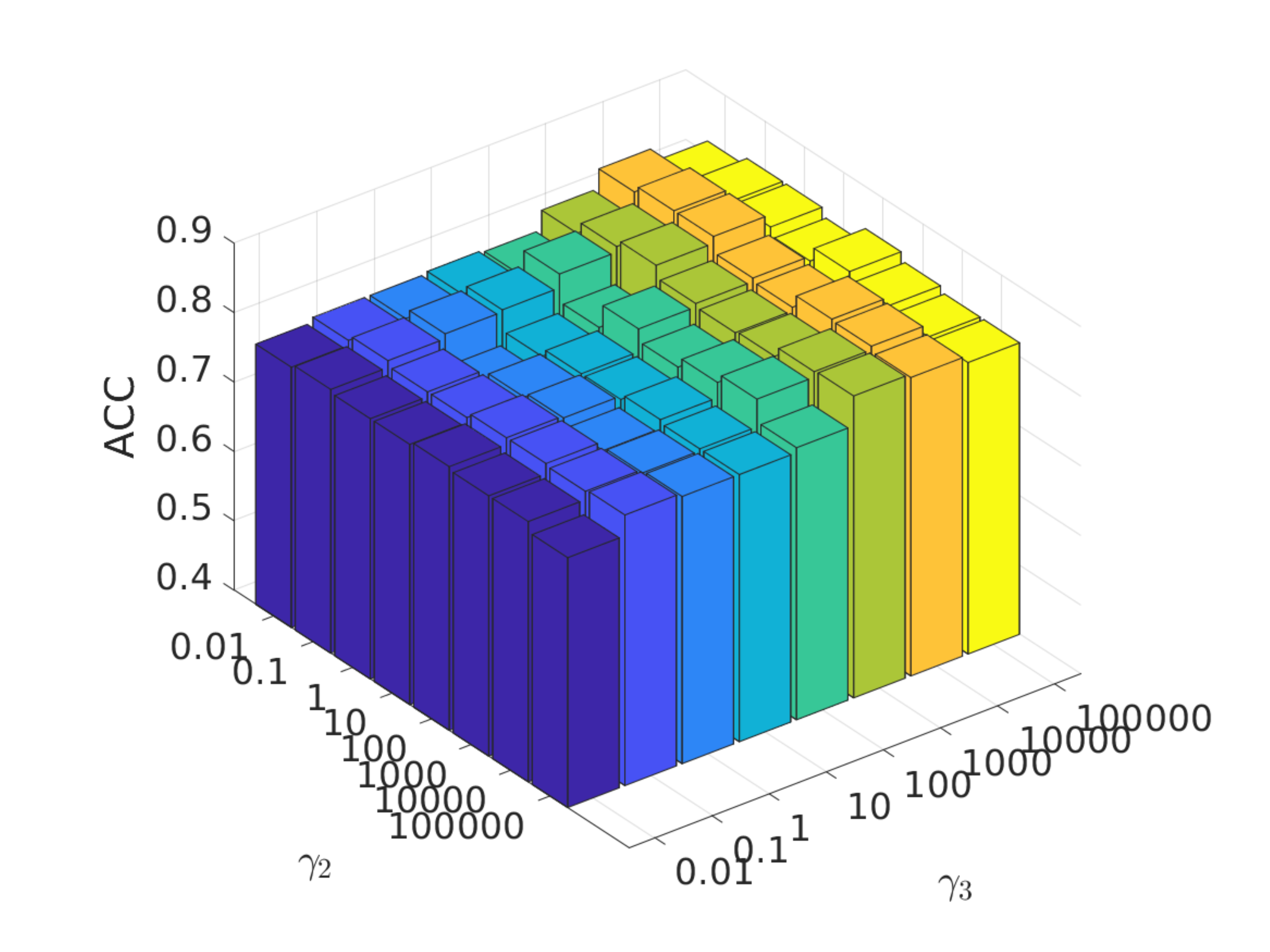}}

\caption{The influence of parameters on ACC of ORL dataset} \label{ORL_acc}
\end{figure*}

\begin{figure*}[!htbp]
\centering
\subfloat[$\gamma_1$=1]{\includegraphics[width=.31\textwidth]{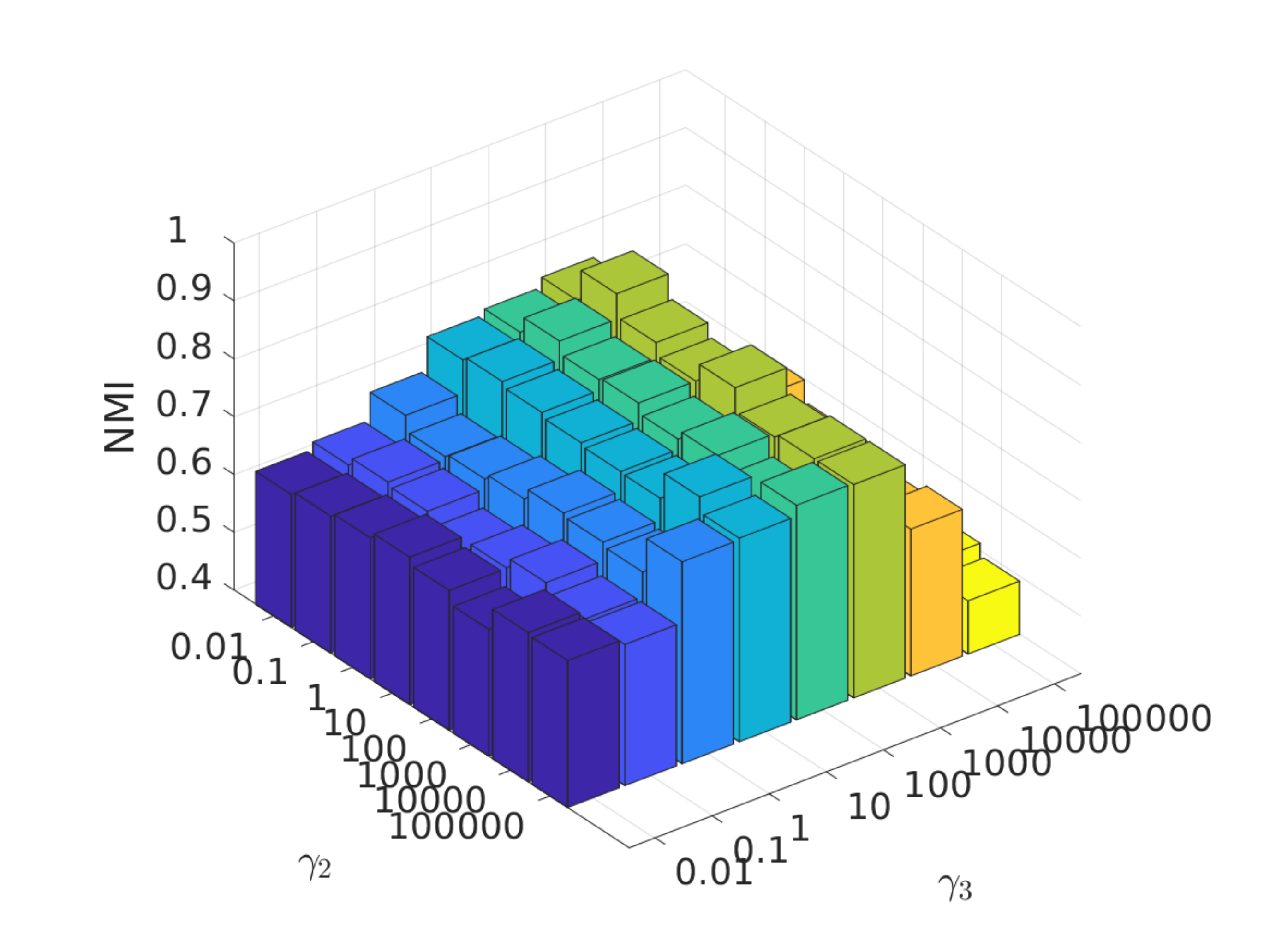}}
\subfloat[$\gamma_1$=1e2]{\includegraphics[width=.31\textwidth]{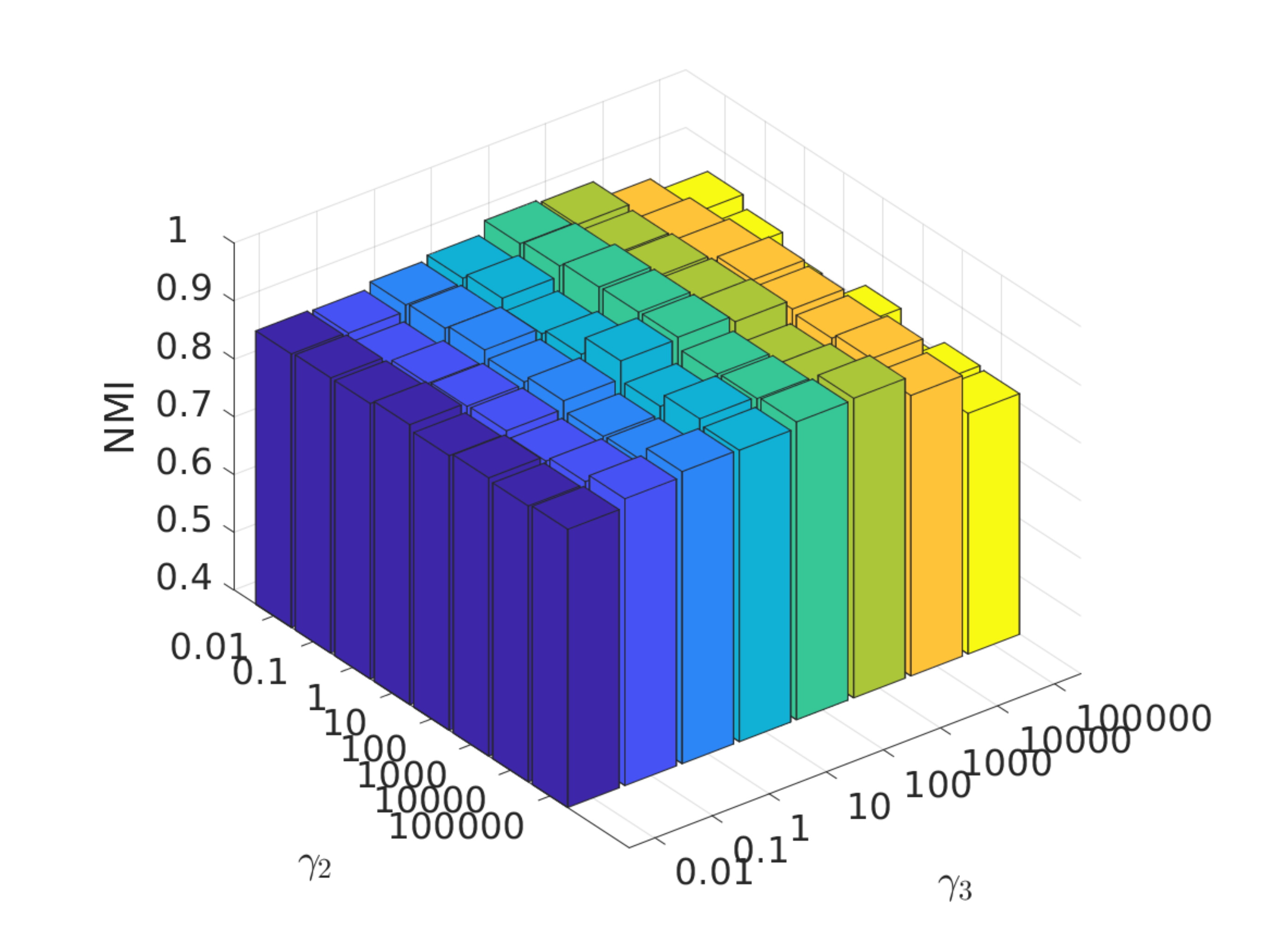}}
\subfloat[$\gamma_1$=1e4]{\includegraphics[width=.31\textwidth]{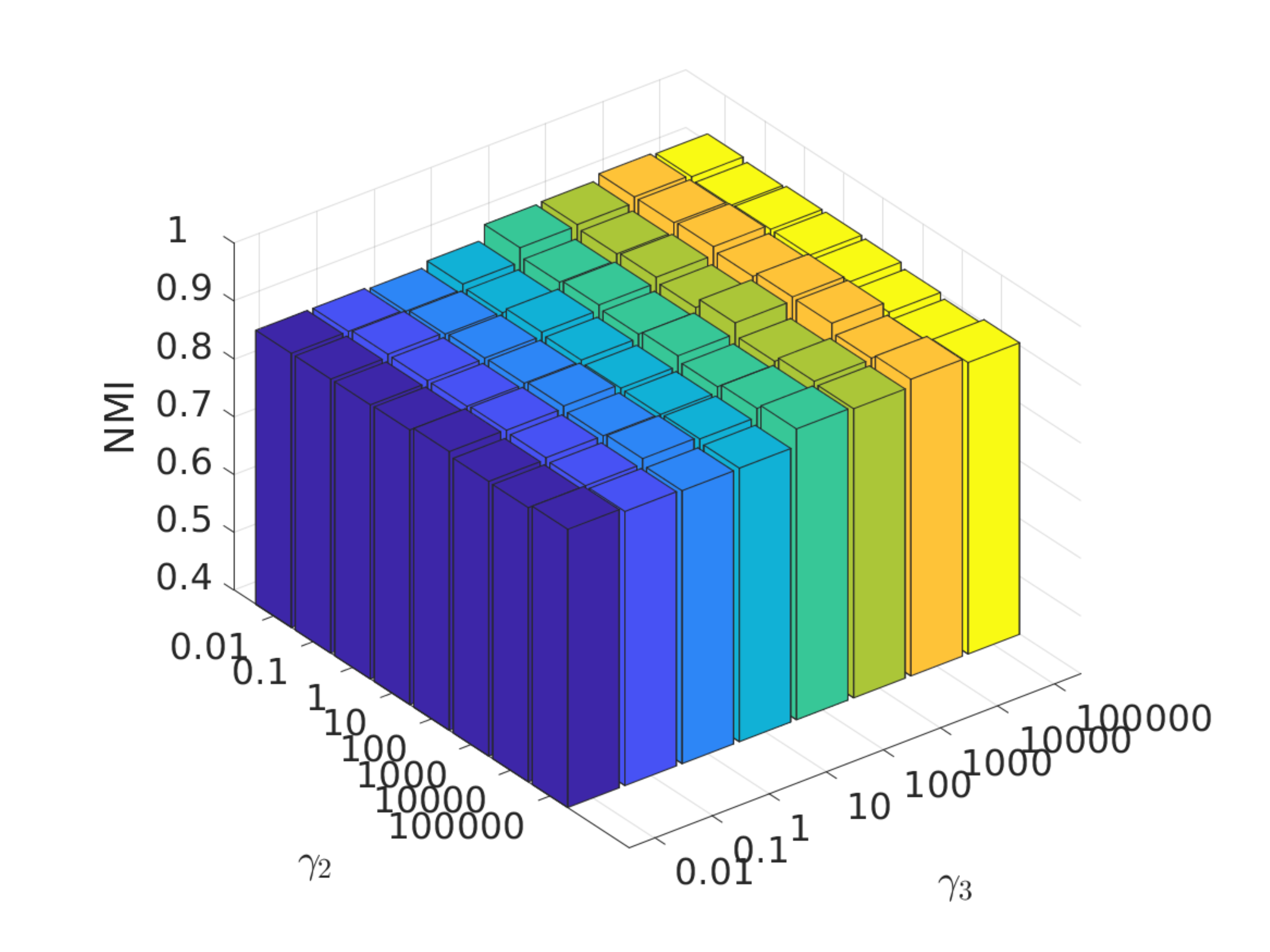}}

\caption{The influence of parameters on NMI of ORL dataset} \label{ORL_nmi}
\end{figure*}

\begin{figure*}[!htbp]
\centering
\subfloat[$\gamma_1$=1e-3]{\includegraphics[width=.31\textwidth]{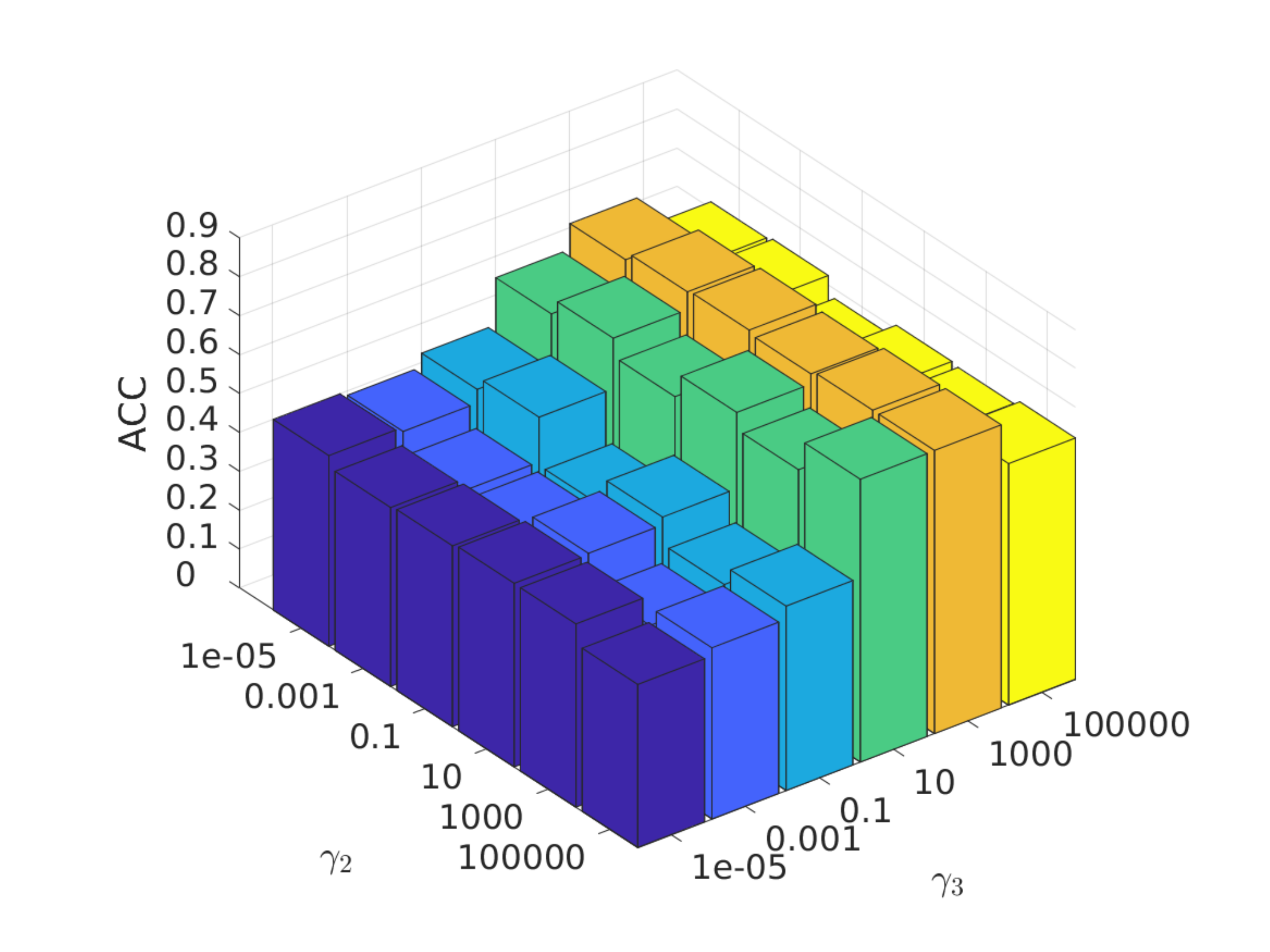}}
\subfloat[$\gamma_1$=1e-2]{\includegraphics[width=.31\textwidth]{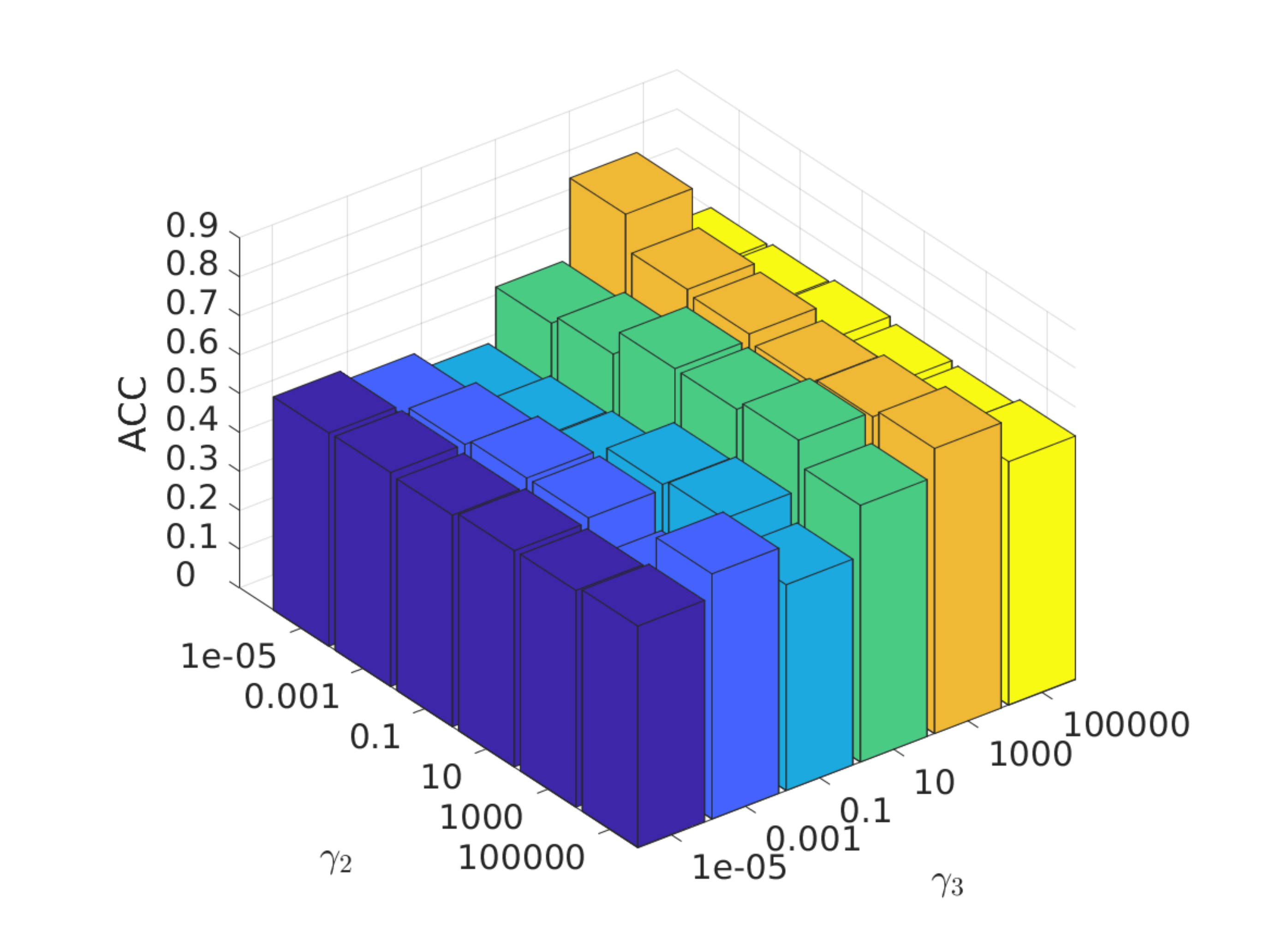}}
\subfloat[$\gamma_1$=1e-1]{\includegraphics[width=.31\textwidth]{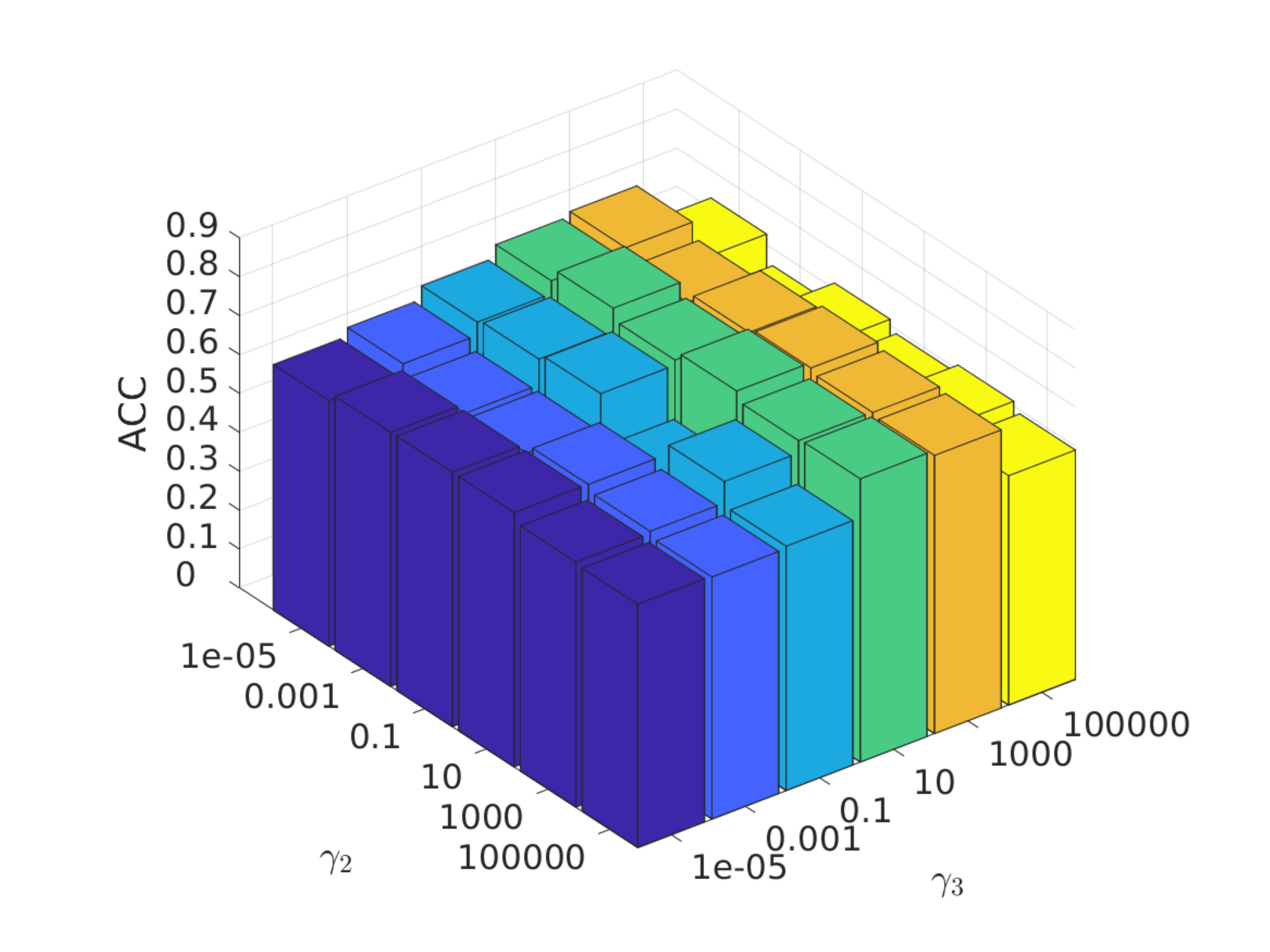}}

\caption{The influence of parameters on ACC of MNIST dataset} \label{MNIST_acc}
\end{figure*}

\begin{figure*}[!htbp]
\centering
\subfloat[$\gamma_1$=1e-3]{\includegraphics[width=.31\textwidth]{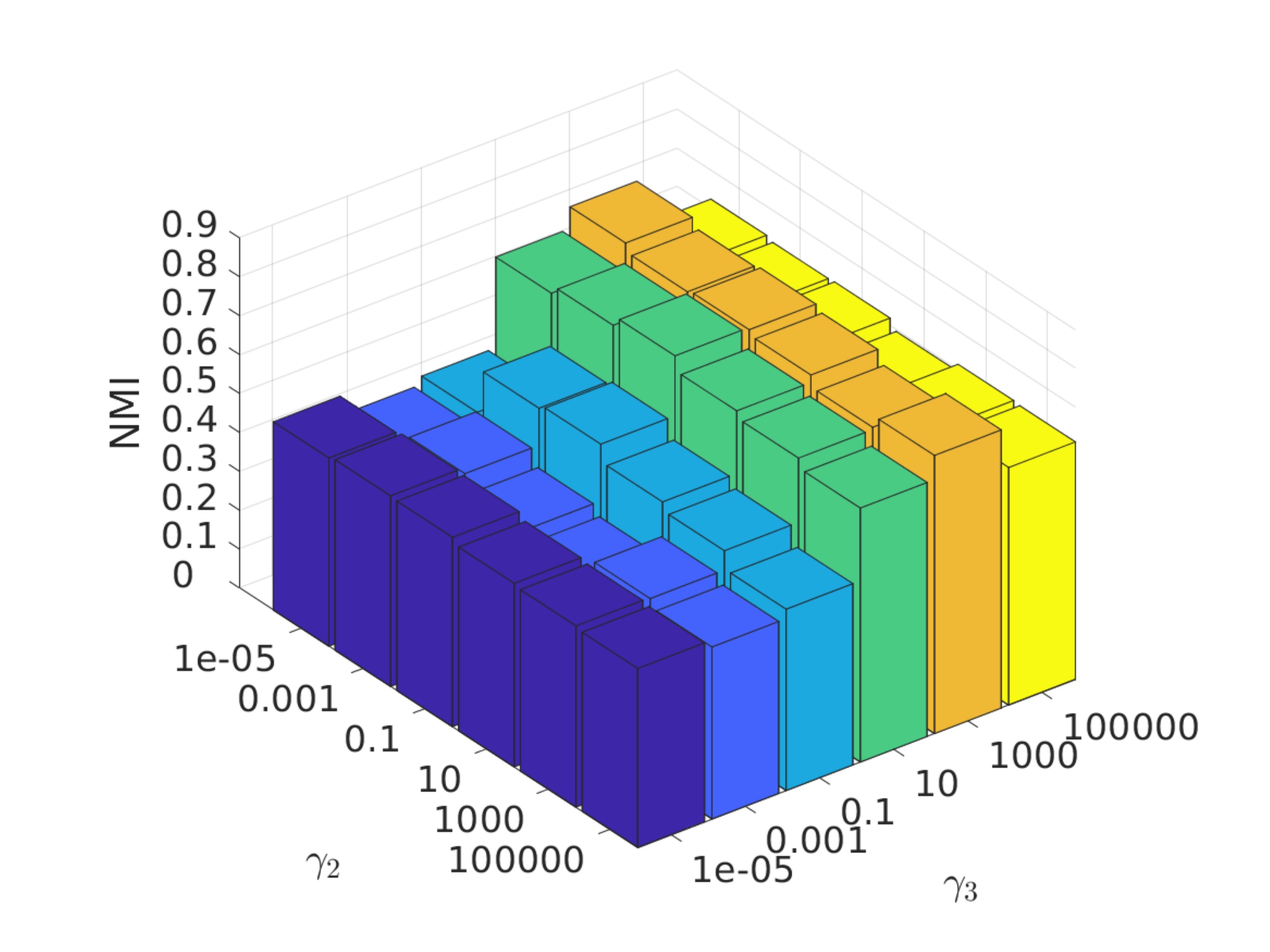}}
\subfloat[$\gamma_1$=1e-2]{\includegraphics[width=.31\textwidth]{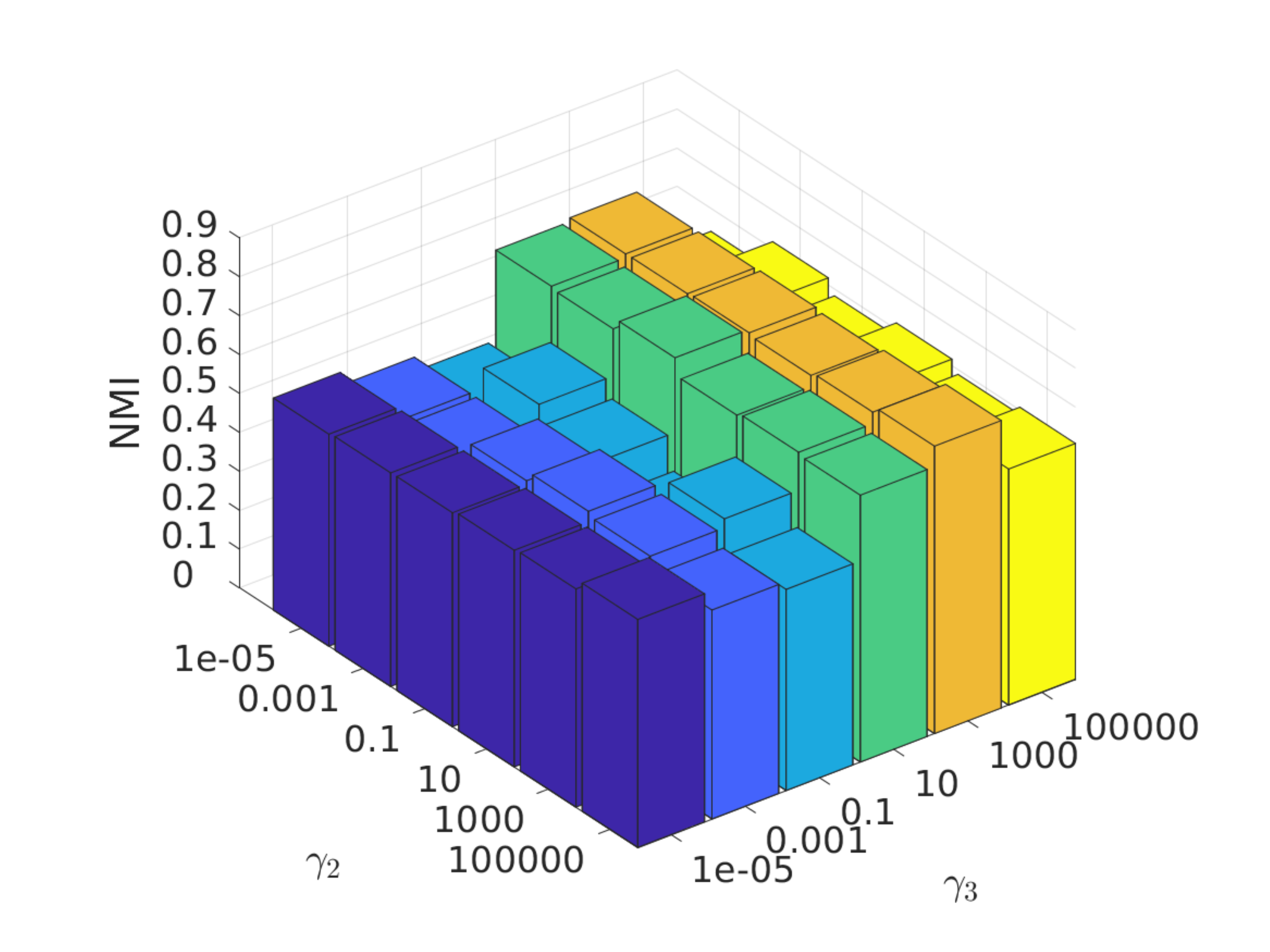}}
\subfloat[$\gamma_1$=1e-1]{\includegraphics[width=.31\textwidth]{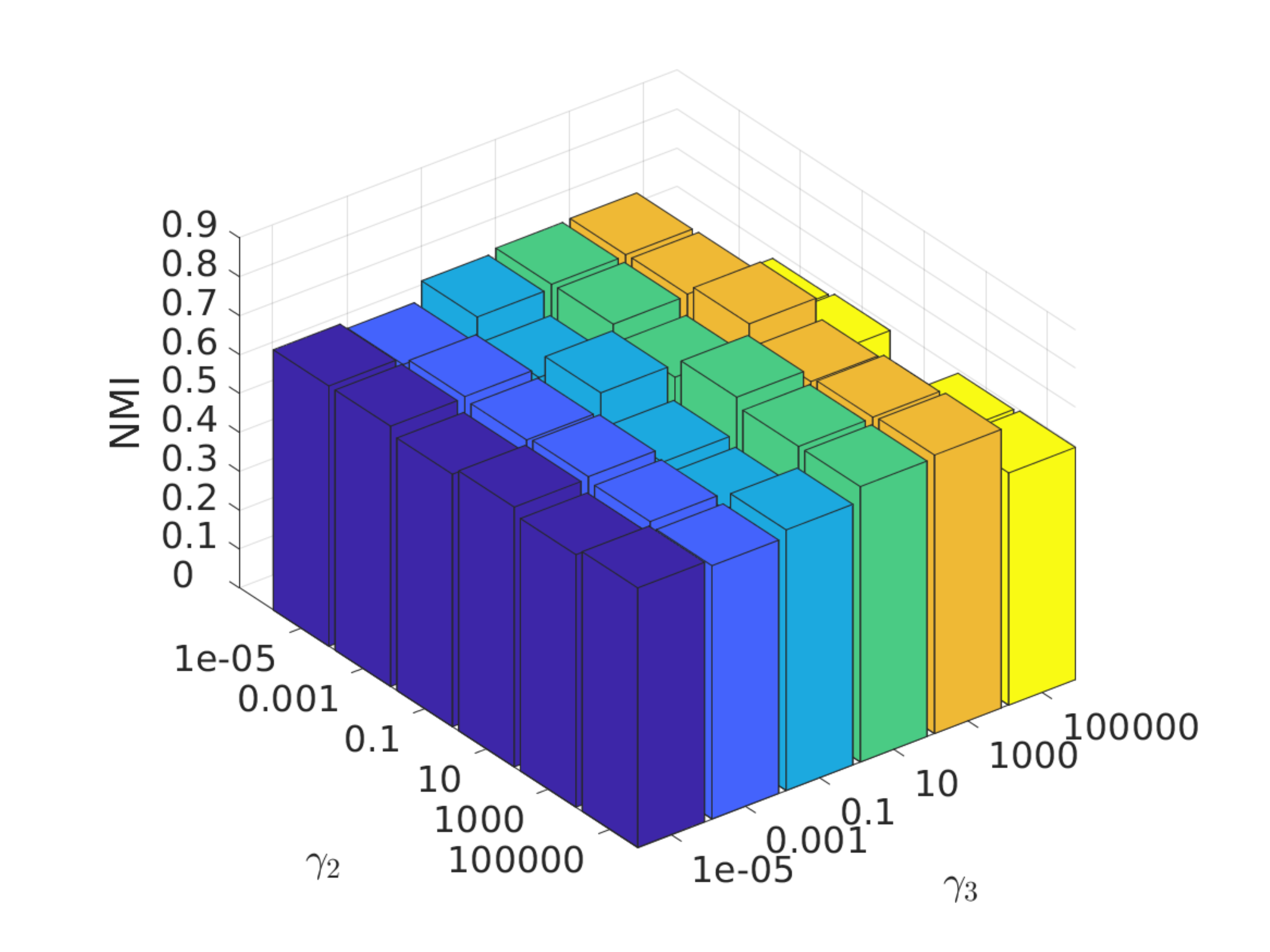}}

\caption{The influence of parameters on NMI of MNIST dataset} \label{MNIST_nmi}
\end{figure*}



\subsection{Parameter Analysis}
There are three hyper-parameters $\gamma_1, \gamma_2, \gamma_3$ in PSSC. Generally, they are data-specific, thus, we perform a grid search to find the best parameter values for each dataset. To demonstrate their impacts on clustering performance, we fix $\gamma_1$ value and show the ACC and NMI variations induced by different $\gamma_2,\gamma_3$ values in Figs.\ref{ORL_acc}-\ref{MNIST_nmi}. Although our method's performance is influenced by the parameters, our method performs well for a wide range of values.

\begin{figure}[!ht]
\centering
\subfloat[$k$=40]{\includegraphics[width=.4\textwidth]{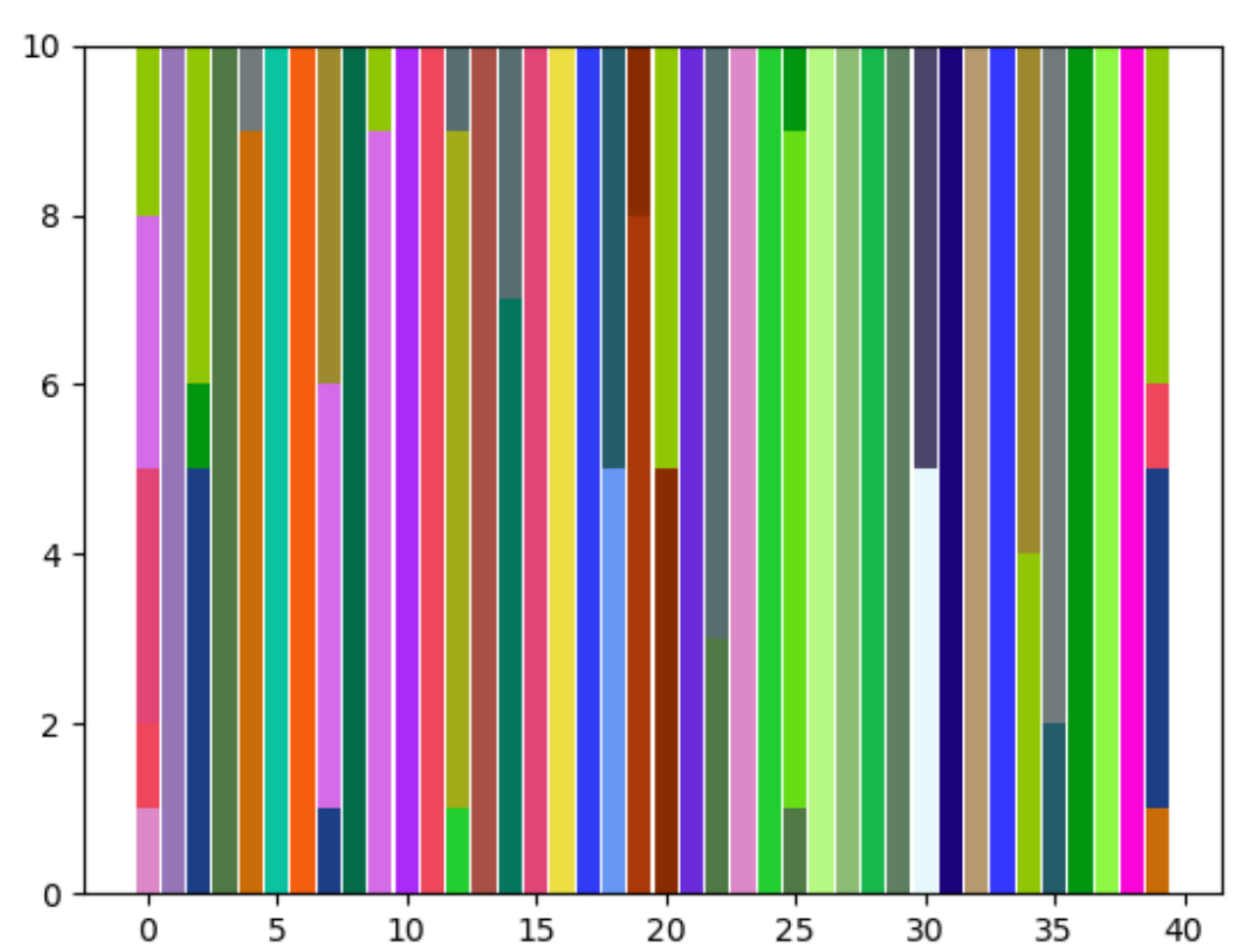}}\\
\subfloat[$k$=45]{\includegraphics[width=.4\textwidth]{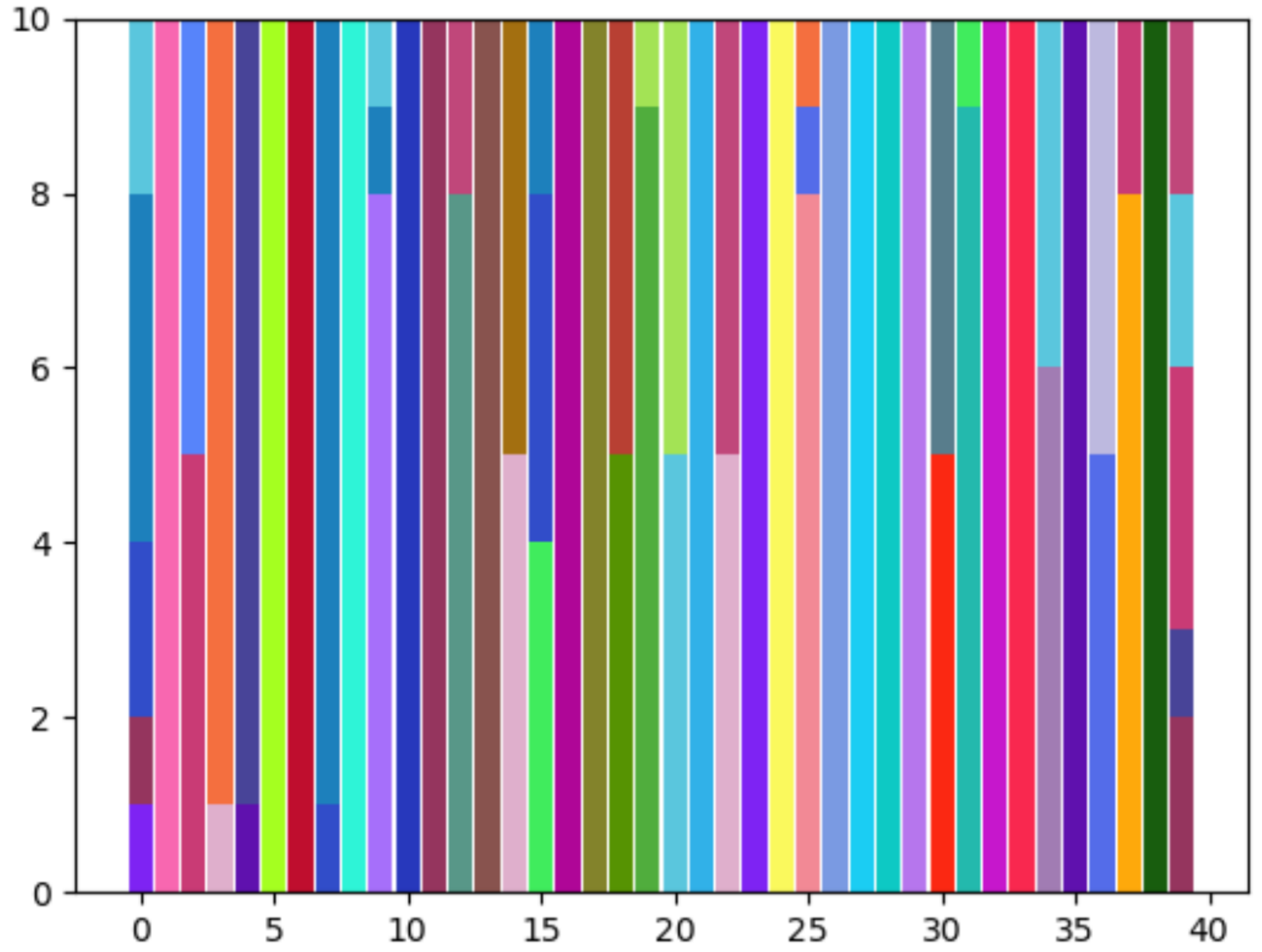}}\\
\subfloat[$k$=50]{\includegraphics[width=.4\textwidth]{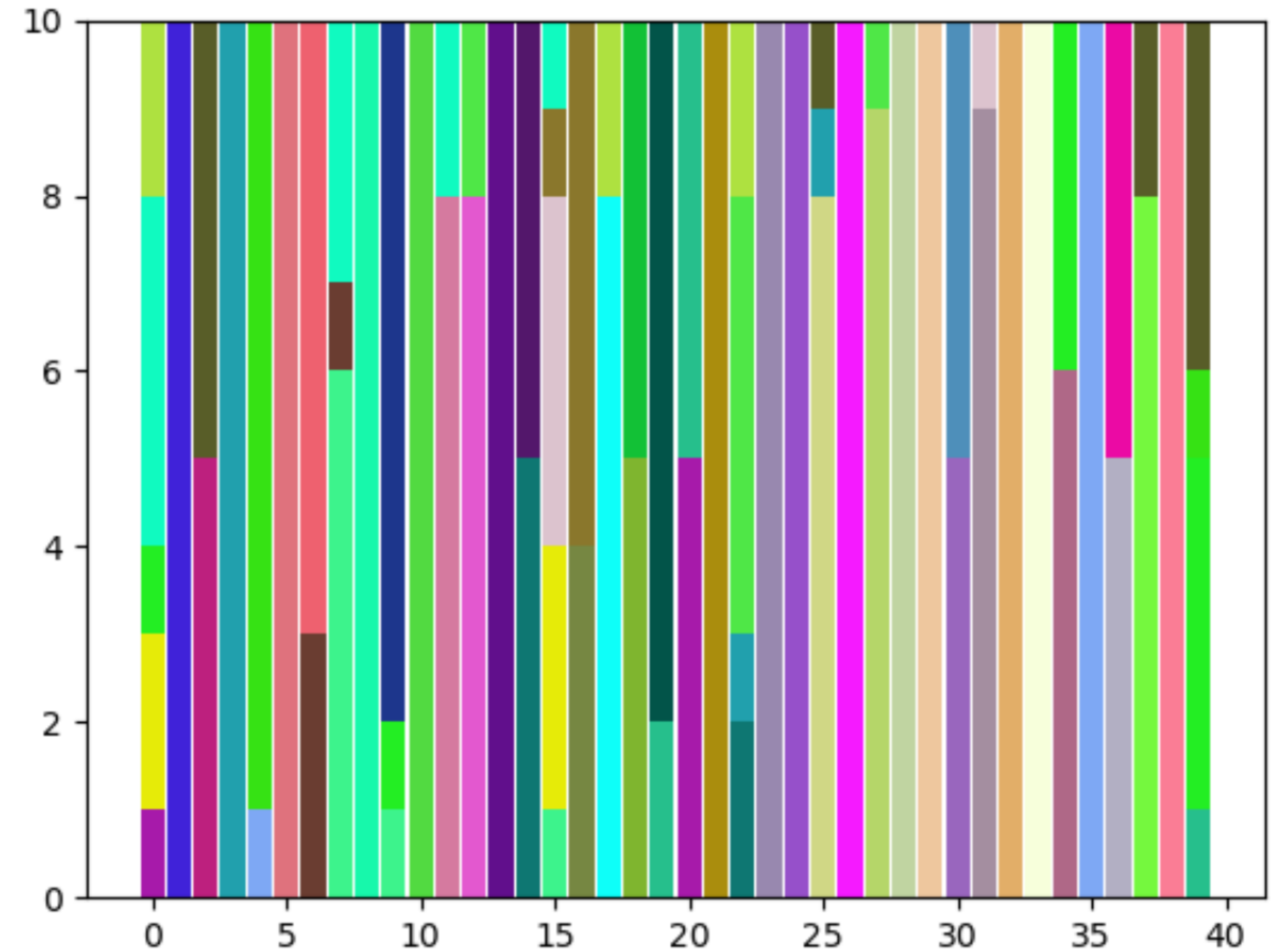}}

\caption{The clustering result based on different cluster number $k$. Every bar in the histogram represents a class in the ORL dataset. Each color denotes a cluster obtained by our method. The height of different bars suggests the quantity of samples.}
\label{number_classes}
\end{figure}

In above analysis, we follow the strategy in clustering community and treat the cluster number $k$ as a priori knowledge. In this part, we change $k$ to find out its impact on performance of ORL dataset. Since ORL has 40 classes, we set $k$ to 45 and 50 to see if no samples get assigned to the additional clusters or different variations within a class. As shown in Fig.\ref{number_classes}, all clusters are assigned with data points even $k$ goes to 50. When $k=40$, there are more bars that only have a single color than other situations, which indicates correct clustering. When $k=50$, more bars are with multiple colors, which indicates that the data samples from one class are split into different clusters.

\begin{figure}[!ht]
\centering
\subfloat[ORL]{\includegraphics[width=.23\textwidth]{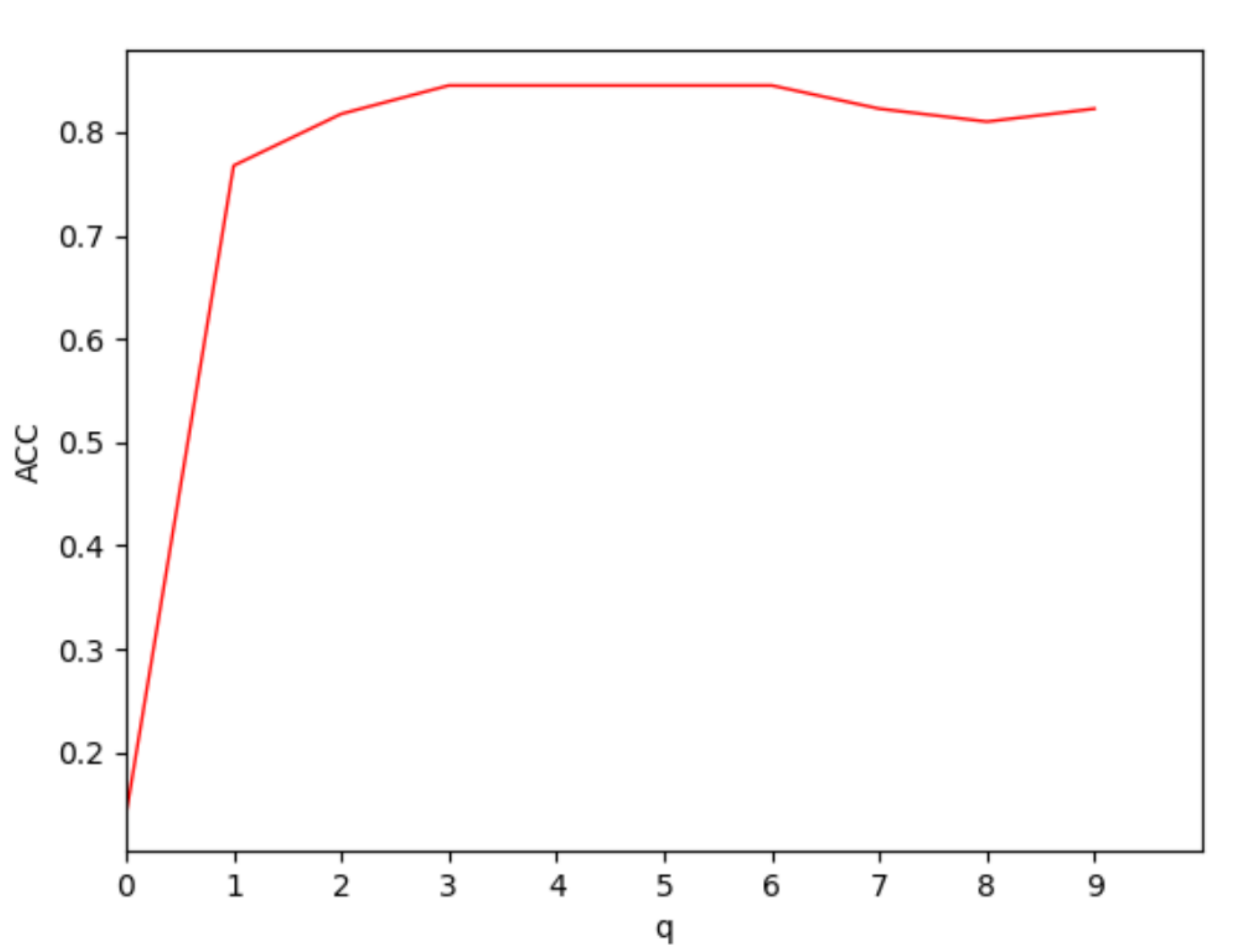}}
\subfloat[COIL20]{\includegraphics[width=.23\textwidth]{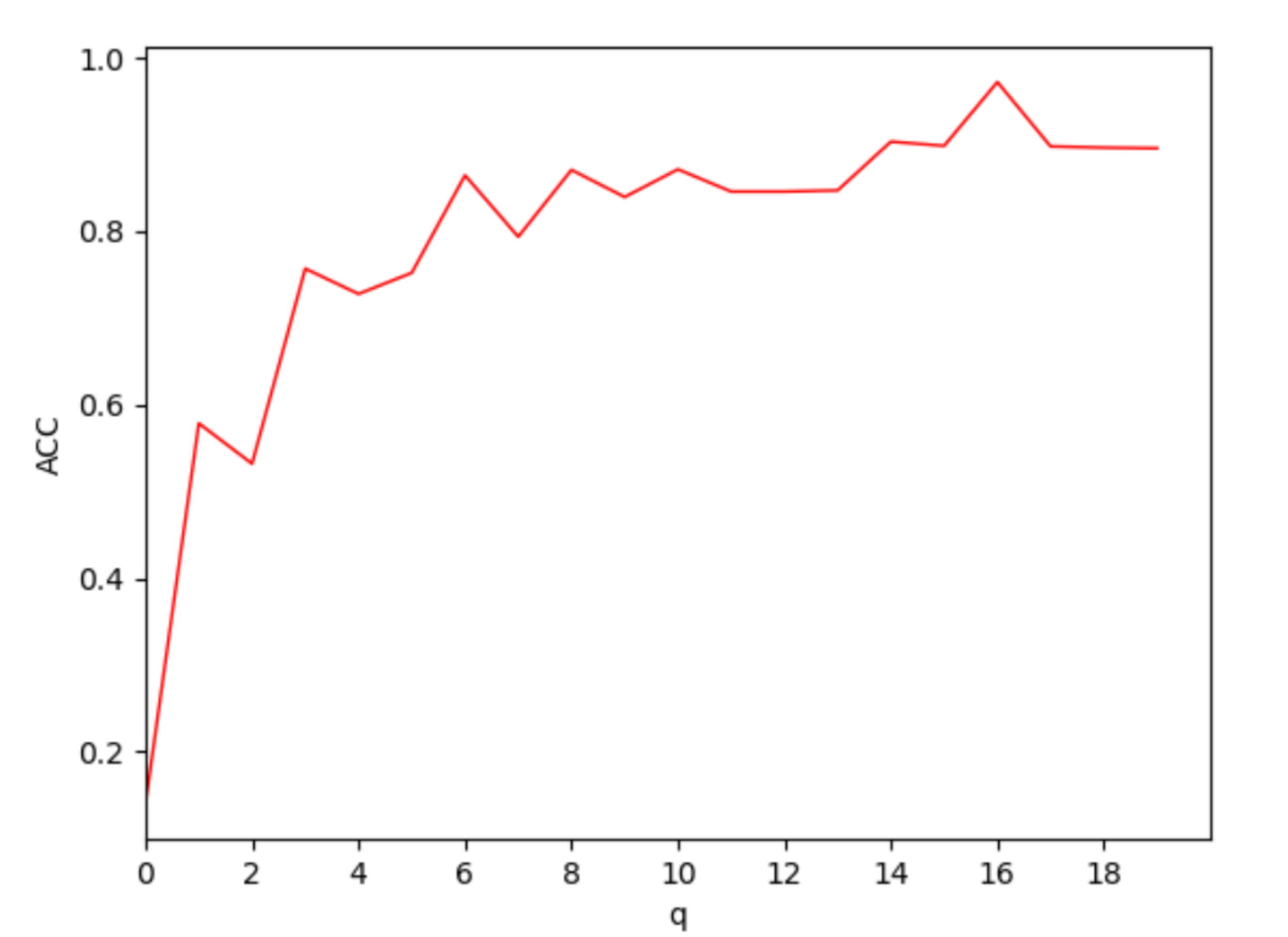}}\\

\caption{The influence of $q$ on ACC of ORL and COIL20 datasets.}
\label{ACC_q}
\end{figure}

As mentioned before, the intrinsic dimension of subspaces $q$ is determined by the data. For example, 3 and 16 are used in DSC for ORL and COIL20 respectively. In Fig.\ref{ACC_q}, we plot the ACC values when $q$ varies. It can be seen that the best performance is achieved when $q$ is the number of nonzero singular values of data points.

\begin{table}{!htbp}
\centering
\renewcommand{\arraystretch}{1.1}
\caption{Statistics of the large datasets.}
\resizebox{.45\textwidth}{!}{
\begin{tabular}{c|r r c}
\hline
Dataset & Samples & Classes & Dimensions \\
\hline
MNIST  & 70,000& 10 & 28$\times$28  \\

USPS   &  9,298 & 10 & 16$\times$16  \\

RCV1   & 10,000& 4  &  2,000  \\
\hline
\end{tabular}}

\label{large_dataset}
\end{table}

\section{Large-scale Experiments}
As mention earlier, the self-expression layer has a size of $n\times n$, which hinders the applications of deep subspace clustering in practice. In this section, we show an approach to mitigate this problem and conduct experiments on large-scale datasets.
\subsection{Datasets}
We perform experiments on two object datasets: MNIST and USPS; a text dataset: RCV1. We use the complete MNIST dataset that has 70,000 handwritten digit images in 10 classes. USPS contains 9,298 handwritten digit images in 10 classes. RCV1 contains approximately 810,000 English news stories labeled with a category tree. Following DKM \cite{fard2018deep}, we randomly sample 10,000 documents from the four largest categories: corporate/industrial, government/social, markets and economics of RCV1, each sample only belongs to one of these four categories. Note that, in contrast to experiments in DEC \cite{xie2016unsupervised} and IDEC \cite{guo2017improved}, we remain samples with multiple labels only if they do not belong to any two of the selected four categories. For text datasets, we select 2,000 words with the highest tf-idf values to represent each document. The statistics of datasets in this experiment are summarized in Table \ref{large_dataset}.

\begin{table}
\centering
\renewcommand{\arraystretch}{1.1}
\caption{Clustering results on MNIST, USPS, and RCV1.}
\resizebox{.45\textwidth}{!}{
\begin{tabular}{|c|c c|c c|c c|}
\hline
Dataset & \multicolumn{2}{c|}{MNIST}  & \multicolumn{2}{c|}{USPS} &  \multicolumn{2}{c|}{RCV1}  \\
\hline
 Methods$\backslash$Metrices& ACC & NMI & ACC & NMI & ACC & NMI \\
 \hline
 
 KM & 0.535& 0.498 & 0.673 & 0.614 & 0.508 & 0.313\\
 \hline

 AE+KM& 0.808 & 0.752  & 0.729  & 0.717  & 0.567 & 0.315 \\
 \hline
  DCN& 0.811  &  0.757  & 0.730   &  0.719  & 0.567 & 0.316  \\
 \hline
 IDEC & 0.867  & 0.864    & 0.752   & 0.749   & 0.595 & 0.347  \\
 \hline
 DKM &  0.840 & 0.796   & 0.757   & 0.776   & 0.583 & 0.331  \\
 \hline
 DCCM  & 0.655  & 0.679   & 0.686   & 0.675   &  - &  - \\
 \hline
 DEPICT & \textbf{0.929} & \textbf{0.879}  & 0.856 &0.865 & - & - \\
  \hline
 DSCDAN & 0.818 & 0.872 & 0.806 & 0.850 & - & -\\
 \hline
 
  $k$SCN & 0.871 & 0.781 & 0.875 & 0.797 & - & - \\
 \hline
 
  PSSC& 0.890  & 0.790  & \textbf{0.935}   &  \textbf{0.856}   &  \textbf{0.780} & \textbf{0.463}  \\
 \hline

\end{tabular}}

\label{large_result}
\end{table}

\subsection{Setup}
\begin{figure}[!htp]
\centering
{\includegraphics[width=.48\textwidth]{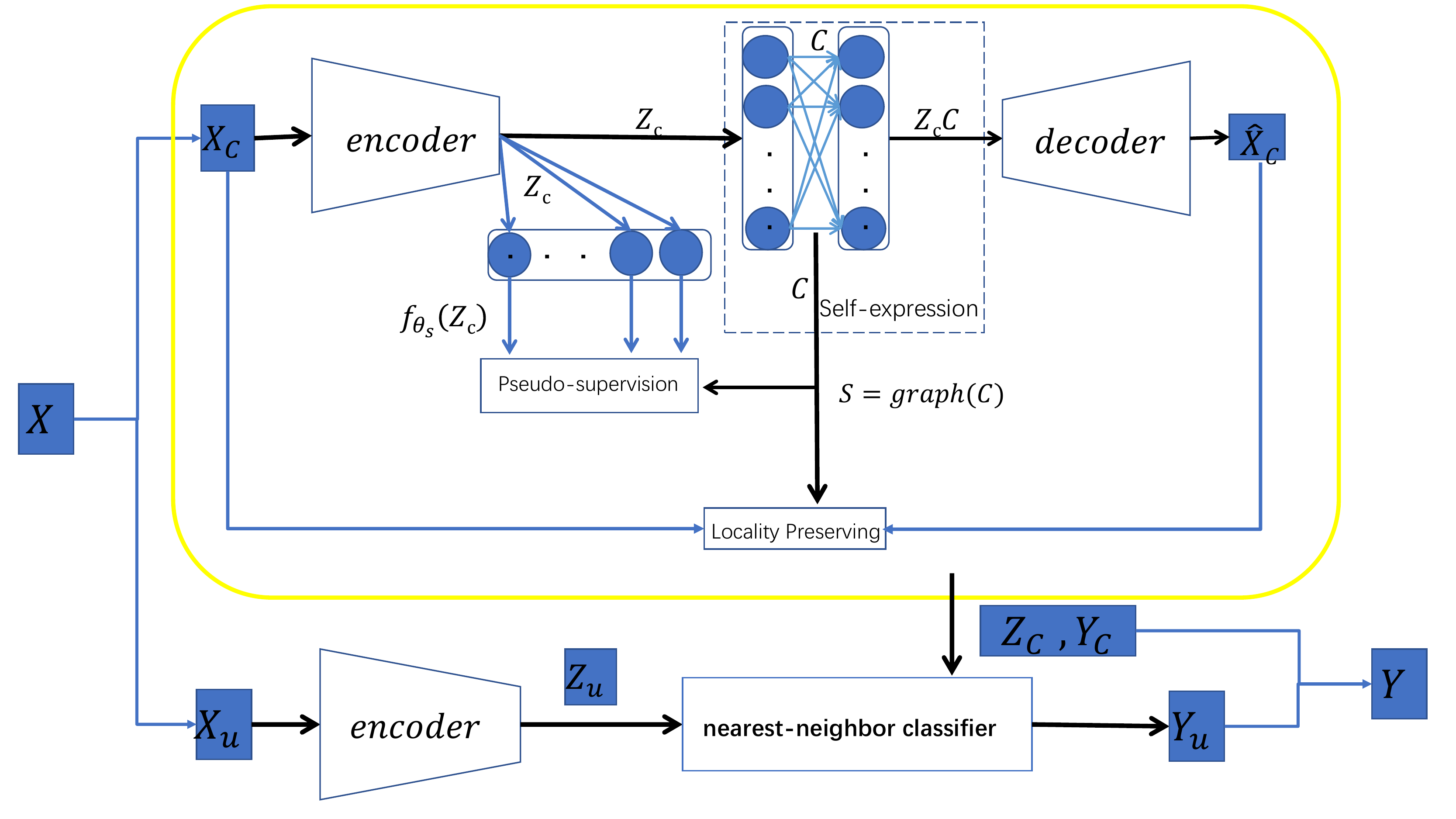}}
\caption{Large-scale experiments are implemented in two steps. First, use a small batch of samples $X_c$ to train PSSC, get the latent representation $Z_c$ and label $Y_c$. Second, encode the rest data $X_u$ to $Z_u$, use nearest-neighbor classifier to obtain label $Y_u$ of the rest data $X_u$. }
\label{large_arch}
\end{figure}

Different from the subspace clustering experiments in the last section, we use latent representations for the clustering task. On the one hand, it can directly demonstrate the quality of our latent representation. On the other hand, it can address out-of-sample and large-scale problems. In particular, we train the network with a reasonable small batch of samples (5,000 samples for each dataset in our experiments); then, we use the similarity matrix to predict the labels of selected samples following the above approach. Finally, we encode all data into latent space and use a nearest-neighbor classifier to predict the labels for the rest of the data. Following this approach, our method can address the out-of-sample problem, which is a long-standing challenge for subspace clustering. The architecture of our method for large-scale experiment is shown in Fig.\ref{large_arch}. The training process is summarized in Algorithm \ref{alg:large}.

\begin{algorithm}
\caption{PSSC for Large-scale Data} 
\label{alg:large} 
\begin{algorithmic}[1]
\REQUIRE
The data set, $X$;\\
The number of clusters, $k$;\\
The size of small batch, $m$;
\ENSURE
The clustering result, $Y$;  
\STATE Randomly sample the training data $X_c$ with $m$ samples, denote the rest as $X_u$;
\STATE Pre-train the auto-encoder;
\STATE Fine-tune PSSC;
\STATE Perform spectral clustering on affinity matrix to obtain label $Y_c$;
\STATE Implement the nearest-neighbor algorithm to obtain the label $Y_u$ of the rest data $X_u$.
\end{algorithmic}
\end{algorithm}

\begin{figure*}[ht]
\centering
\subfloat[IDEC\label{IDEC-v}]{\includegraphics[width=.3\textwidth]{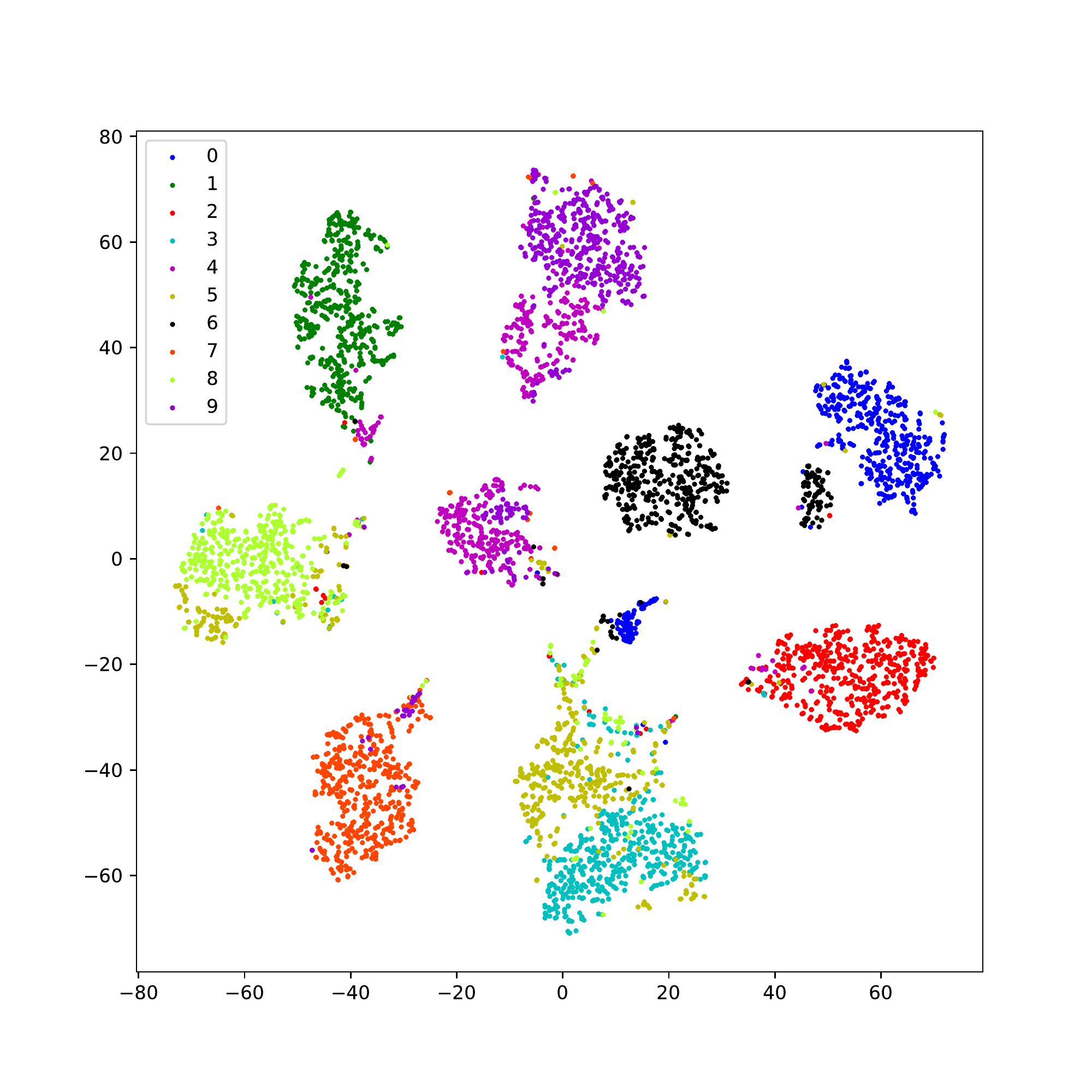}}
\subfloat[DKM\label{DKM-v}]{\includegraphics[width=.3\textwidth]{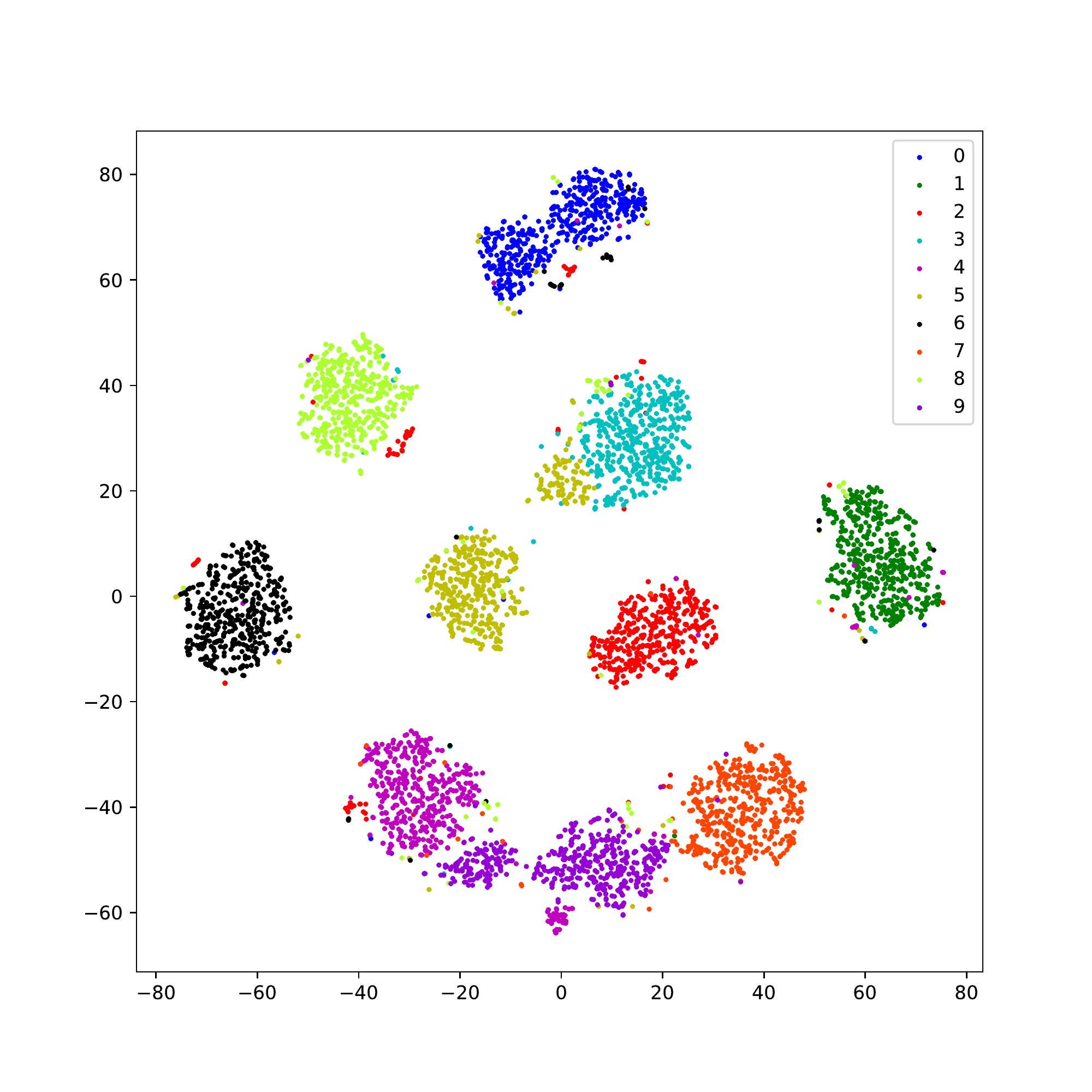}}
\subfloat[DEPICT\label{DEPICT-v}]{\includegraphics[width=.3\textwidth]{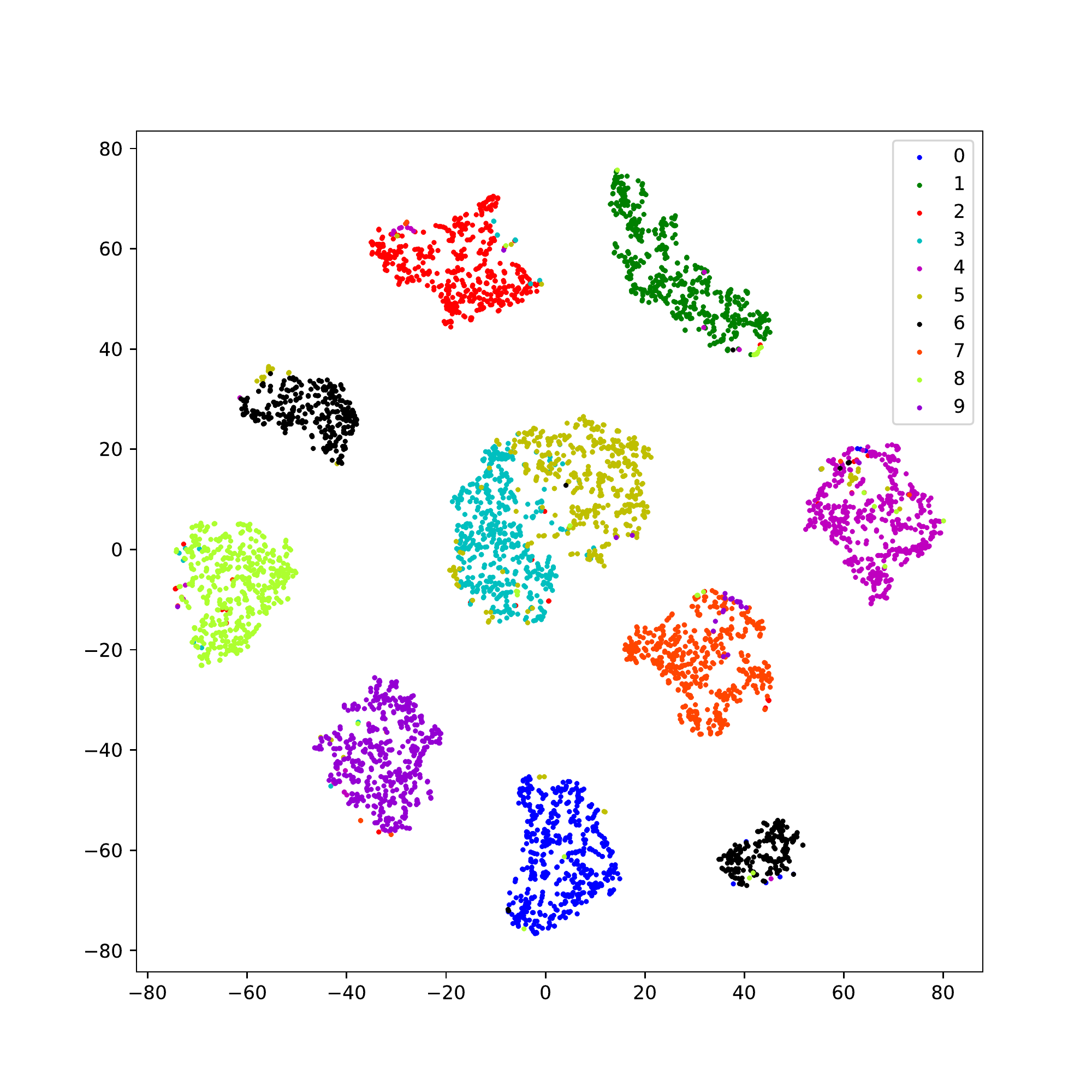}}\\
\subfloat[DSCDAN\label{DSCDAN-v}]{\includegraphics[width=.3\textwidth]{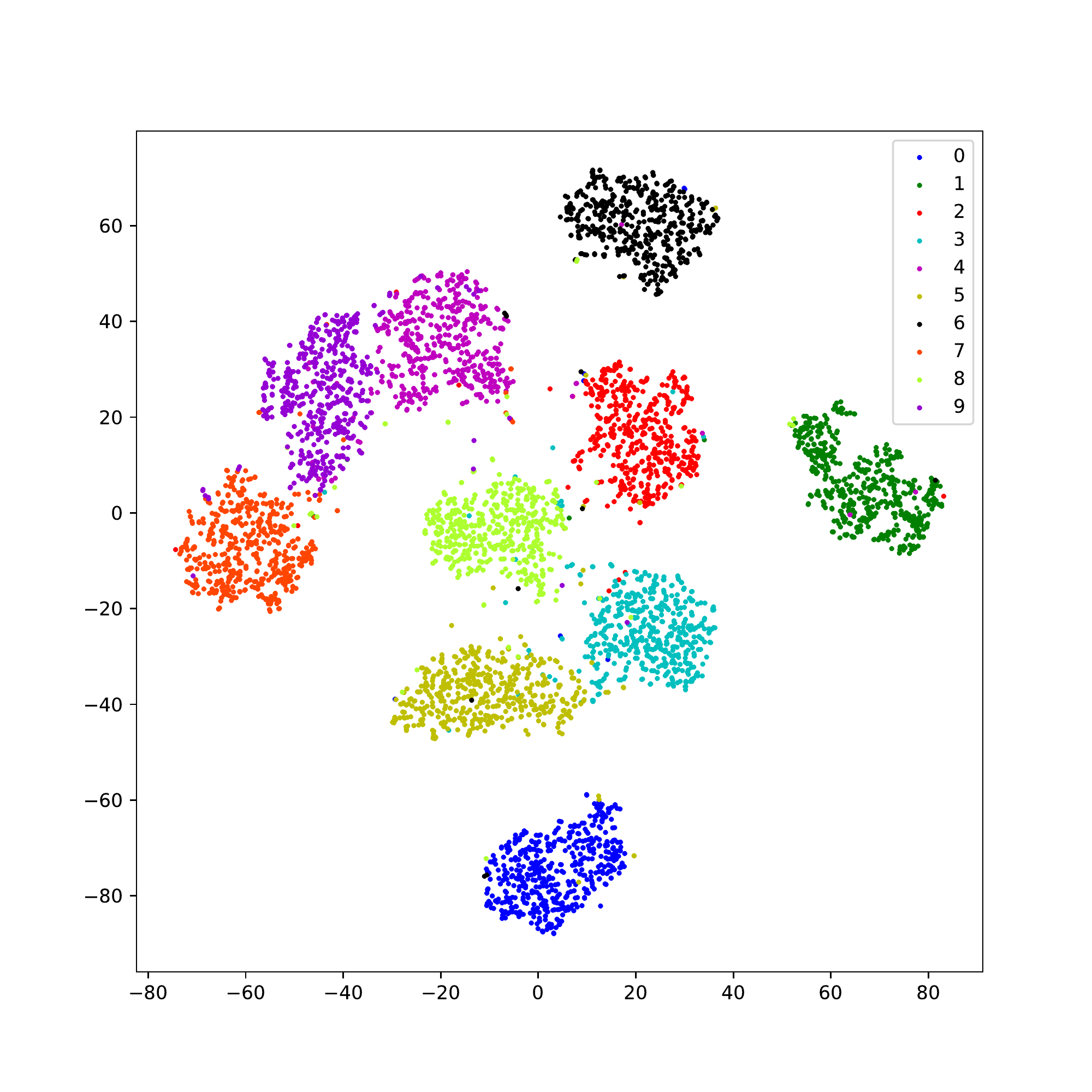}}
\subfloat[$k$SCN\label{kscn-v}]{\includegraphics[width=.3\textwidth]{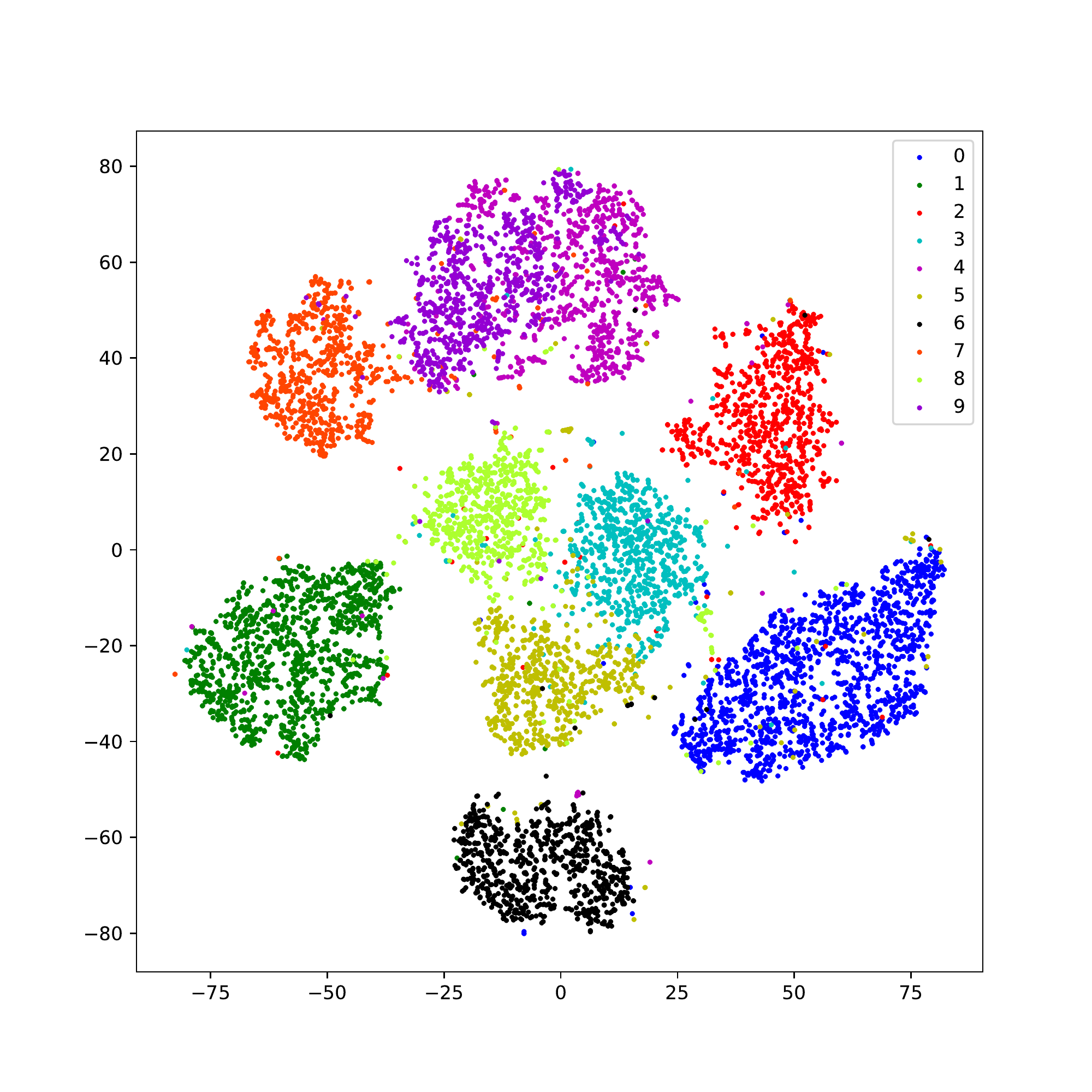}}
\subfloat[PSSC\label{SAE-v}]{\includegraphics[width=.3\textwidth]{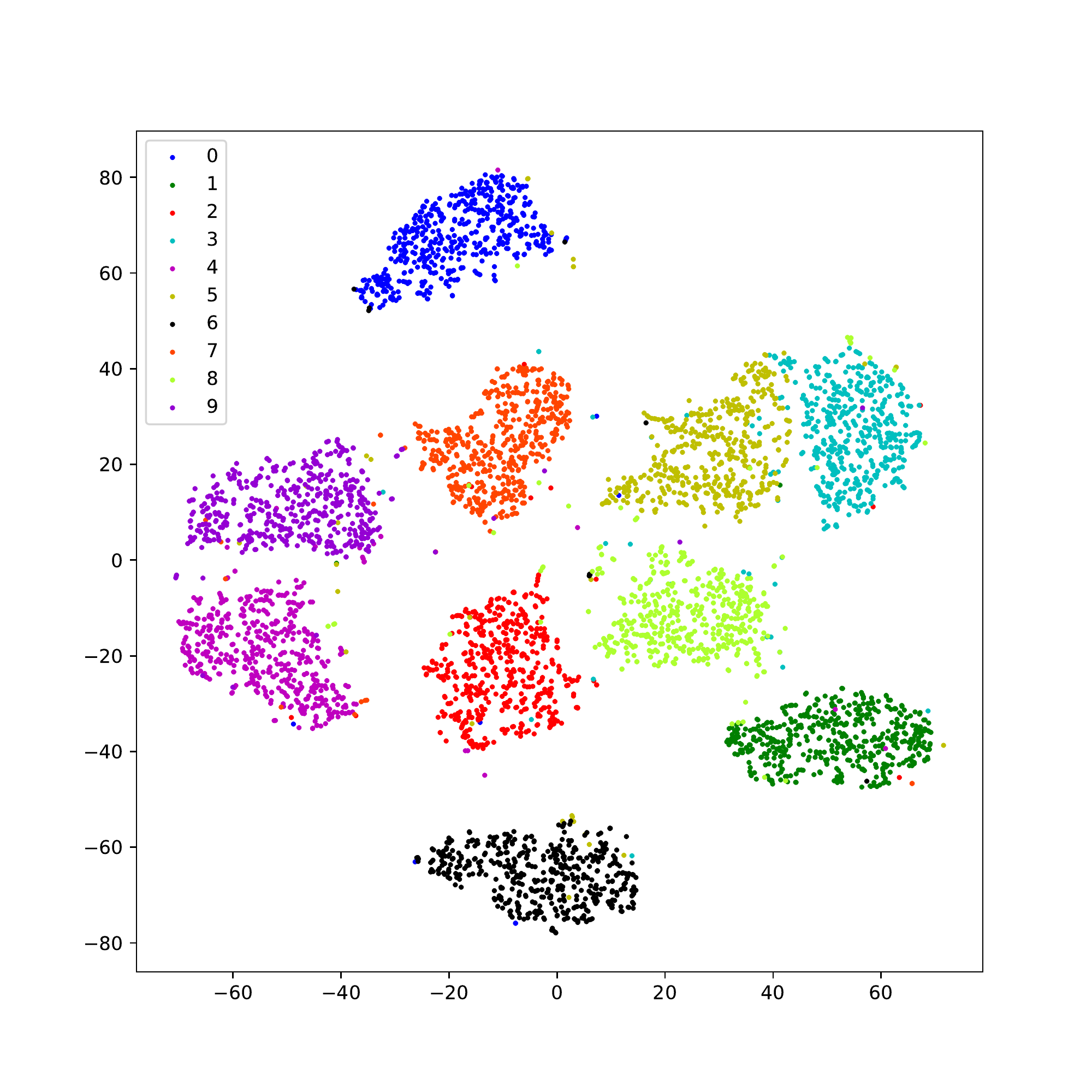}}
\caption{2D visualization of the embedding spaces learned on USPS dataset.}
\label{tsne}
\end{figure*}

For a fair comparison, we use the same encoder/decoder architecture as DEC \cite{xie2016unsupervised}, IDEC \cite{guo2017improved}, and DKM \cite{fard2018deep}. The encoder is a fully-connected network with dimensions of $d$-500-500-2000-$k$ for all datasets, where $d$ is the dimension of input features and $k$ is the number of clusters. Correspondingly, the decoder is a mirror of the encoder, a fully-connected network with dimensions of $k$-2000-500-500-$d$. A ReLU activation function is used for each layer except the input, output, and self-expression layers. We pre-train the AE in 50 epochs and fine-tune the whole network with an objective function (\ref{objf}) in 30 epochs.

We compare PSSC with $k$-means clustering (KM), an AE with KM applied to latent representation (AE+KM), and recent deep clustering approaches: DCN \cite{yang2017towards}, IDEC \cite{guo2017improved}, DKM \cite{fard2018deep}, DCCM \cite{wu2019deep}, DEPICT \cite{ghasedi2017deep}, DSCDAN \cite{yang2019deep}, and $k$SCN \cite{zhang2018scalable}. Note that we did not use DCCM,  DEPICT, $k$SCN, and DSCDAN on the text data because they are designed for image datasets. We use the SVD version of it because it gets the best performance in \cite{zhang2018scalable}.

\subsection{Results}
Table \ref{large_result} summarizes the clustering performance of the above methods on three datasets. For each measure, the best result is highlighted in boldface. It can be observed that our method still outperforms recent deep clustering methods. In particular, compared to recent DKM, PSSC improves ACC by 5\%, 17.8\%, and 19.7\% on the three datasets, respectively. With regard to DCCM, ACC improvement is more than 23\%. Although the ACC of DEPICT on MNIST is approximately 3.9\% higher than our method, our method achieves 7.9\% higher on USPS. This performance beneﬁts from embedding samples into subspaces and pseudo-supervising the learning process. $k$SCN is designed for deep subspace clustering on large image datasets. Compared to it, our method improves ACC by 2\% and 6\% on MNIST and USPS, respectively. Our method shows a convincing ability to address large-scale and out-of-sample problems.

Using USPS data as an example, we use the t-SNE \cite{maaten2008visualizing}
method to visualize the learned latent representations of several methods that produce good performance. As shown in Fig.\ref{tsne}, samples of different classes merge in IDEC and DKM, which is because they force samples to move to cluster centers. 
In DEPICT, the black class is separated into a different place, which could degrade the performance. DSCDSN performs a little better but still has a problem that several clusters are close to each other. The same problem bothers $k$SCN, the pink and the purple classes blend with each other. Our method can precisely project samples of each class into a subspace, thus, samples can be well separated.

\section{Conclusion}
In this paper, we propose a novel clustering method, named PSSC, by combining DSC and the self-supervised learning approach. The method uses pseudo-labels as supervision and makes full use of relation information to learn discriminative features. To the best of our knowledge, it is the ﬁrst effort that integrates locality preserving, self-expression, and self-supervision into a unified framework. Extensive experimental results on both small-scale and large-scale datasets have shown the superiority of the proposed method on similarity and representation learning over state-of-the-art methods, including the latest deep learning based techniques. In particular, we demonstrate how to address the large-scale and out-of-sample problem facing subspace clustering. In future work, it is interesting to explore different self-supervision approaches and further investigate the cluster structure of complex sample distributions.

\bibliographystyle{IEEEtran}
\bibliography{ref}

\begin{thebibliography}{10}
\providecommand{\url}[1]{#1}
\csname url@samestyle\endcsname
\providecommand{\newblock}{\relax}
\providecommand{\bibinfo}[2]{#2}
\providecommand{\BIBentrySTDinterwordspacing}{\spaceskip=0pt\relax}
\providecommand{\BIBentryALTinterwordstretchfactor}{4}
\providecommand{\BIBentryALTinterwordspacing}{\spaceskip=\fontdimen2\font plus
\BIBentryALTinterwordstretchfactor\fontdimen3\font minus
  \fontdimen4\font\relax}
\providecommand{\BIBforeignlanguage}[2]{{%
\expandafter\ifx\csname l@#1\endcsname\relax
\typeout{** WARNING: IEEEtran.bst: No hyphenation pattern has been}%
\typeout{** loaded for the language `#1'. Using the pattern for}%
\typeout{** the default language instead.}%
\else
\language=\csname l@#1\endcsname
\fi
#2}}
\providecommand{\BIBdecl}{\relax}
\BIBdecl

\bibitem{jain2010data}
A.~K. Jain, ``Data clustering: 50 years beyond k-means,'' \emph{Pattern
  recognition letters}, vol.~31, no.~8, pp. 651--666, 2010.

\bibitem{johnson1967hierarchical}
S.~C. Johnson, ``Hierarchical clustering schemes,'' \emph{Psychometrika},
  vol.~32, no.~3, pp. 241--254, 1967.

\bibitem{ng2002spectral}
A.~Y. Ng, M.~I. Jordan, and Y.~Weiss, ``On spectral clustering: Analysis and an
  algorithm,'' in \emph{Advances in neural information processing systems},
  2002, pp. 849--856.

\bibitem{agrawal1998automatic}
R.~Agrawal, J.~Gehrke, D.~Gunopulos, and P.~Raghavan, ``Automatic subspace
  clustering of high dimensional data for data mining applications,'' in
  \emph{Proceedings of the 1998 ACM SIGMOD international conference on
  Management of data}, 1998, pp. 94--105.

\bibitem{elhamifar2013sparse}
E.~Elhamifar and R.~Vidal, ``Sparse subspace clustering: Algorithm, theory, and
  applications,'' \emph{IEEE transactions on pattern analysis and machine
  intelligence}, vol.~35, no.~11, pp. 2765--2781, 2013.

\bibitem{liu2012robust}
G.~Liu, Z.~Lin, S.~Yan, J.~Sun, Y.~Yu, and Y.~Ma, ``Robust recovery of subspace
  structures by low-rank representation,'' \emph{IEEE transactions on pattern
  analysis and machine intelligence}, vol.~35, no.~1, pp. 171--184, 2012.

\bibitem{tang2020cgd}
C.~Tang, X.~Liu, X.~Zhu, E.~Zhu, Z.~Luo, L.~Wang, and W.~Gao, ``Cgd: Multi-view
  clustering via cross-view graph diffusion.'' in \emph{AAAI Conference on
  Artificial Intelligence}, 2020, pp. 5924--5931.

\bibitem{lv2021multi}
J.~Lv, Z.~Kang, B.~Wang, L.~Ji, and Z.~Xu, ``Multi-view subspace clustering via
  partition fusion,'' \emph{Information Sciences}, vol. 560, pp. 410--423,
  2021.

\bibitem{peng2016constructing}
X.~Peng, Z.~Yu, Z.~Yi, and H.~Tang, ``Constructing the l2-graph for robust
  subspace learning and subspace clustering,'' \emph{IEEE transactions on
  cybernetics}, vol.~47, no.~4, pp. 1053--1066, 2016.

\bibitem{TangTMM2018}
C.~Tang, X.~Zhu, X.~Liu, M.~Li, P.~Wang, C.~Zhang, and L.~Wang, ``Learning a
  joint affinity graph for multiview subspace clustering,'' \emph{IEEE
  Transactions on Multimedia}, vol.~21, no.~7, pp. 1724--1736, 2019.

\bibitem{Cao2015Constrained}
X.~Cao, C.~Zhang, C.~Zhou, H.~Fu, and H.~Foroosh, ``Constrained multi-view
  video face clustering,'' \emph{IEEE Transactions on Image Processing},
  vol.~24, no.~11, pp. 4381--4393, 2015.

\bibitem{Yin2018Subspace}
M.~Yin, S.~Xie, Z.~Wu, Y.~Zhang, and J.~Gao, ``Subspace clustering via learning
  an adaptive low-rank graph,'' \emph{IEEE Transactions on Image Processing A
  Publication of the IEEE Signal Processing Society}, pp. 1--1, 2018.

\bibitem{vidal2014low}
R.~Vidal and P.~Favaro, ``Low rank subspace clustering (lrsc),'' \emph{Pattern
  Recognition Letters}, vol.~43, pp. 47--61, 2014.

\bibitem{TangTKDE2019}
C.~Tang, X.~Liu, X.~Zhu, J.~Xiong, M.~Li, J.~Xia, X.~Wang, and L.~Wang,
  ``Feature selective projection with low-rank embedding and dual laplacian
  regularization,'' \emph{IEEE Transactions on Knowledge and Data Engineering},
  pp. 1--1, 2019.

\bibitem{li2015structured}
C.-G. Li and R.~Vidal, ``Structured sparse subspace clustering: A unified
  optimization framework,'' in \emph{Proceedings of the IEEE conference on
  computer vision and pattern recognition}, 2015, pp. 277--286.

\bibitem{Zhenwen2019Learning}
Z.~Ren, Q.~Sun, B.~Wu, X.~Zhang, and W.~Yan, ``Learning latent low-rank and
  sparse embedding for robust image feature extraction,'' \emph{IEEE
  Transactions on Image Processing}, vol.~PP, no.~99, pp. 1--1, 2019.

\bibitem{lu2012robust}
C.-Y. Lu, H.~Min, Z.-Q. Zhao, L.~Zhu, D.-S. Huang, and S.~Yan, ``Robust and
  efficient subspace segmentation via least squares regression,'' in
  \emph{European conference on computer vision}.\hskip 1em plus 0.5em minus
  0.4em\relax Springer, 2012, pp. 347--360.

\bibitem{peng2015robust}
X.~Peng, Z.~Yi, and H.~Tang, ``Robust subspace clustering via thresholding
  ridge regression,'' in \emph{Twenty-Ninth AAAI Conference on Artificial
  Intelligence}, 2015.

\bibitem{kang2021structured}
Z.~Kang, Z.~Lin, X.~Zhu, and W.~Xu, ``Structured graph learning for scalable
  subspace clustering: From single-view to multi-view,'' \emph{IEEE
  Transactions on Cybernetics}, 2021.

\bibitem{zhan2017graph}
K.~Zhan, C.~Zhang, J.~Guan, and J.~Wang, ``Graph learning for multiview
  clustering,'' \emph{IEEE transactions on cybernetics}, vol.~48, no.~10, pp.
  2887--2895, 2017.

\bibitem{patel2015latent}
V.~M. Patel, H.~Van~Nguyen, and R.~Vidal, ``Latent space sparse and low-rank
  subspace clustering,'' \emph{IEEE Journal of Selected Topics in Signal
  Processing}, vol.~9, no.~4, pp. 691--701, 2015.

\bibitem{zhang2018generalized}
C.~Zhang, H.~Fu, Q.~Hu, X.~Cao, Y.~Xie, D.~Tao, and D.~Xu, ``Generalized latent
  multi-view subspace clustering,'' \emph{IEEE transactions on pattern analysis
  and machine intelligence}, vol.~42, no.~1, pp. 86--99, 2018.

\bibitem{yin2016kernel}
M.~Yin, Y.~Guo, J.~Gao, Z.~He, and S.~Xie, ``Kernel sparse subspace clustering
  on symmetric positive definite manifolds,'' in \emph{Proceedings of the IEEE
  Conference on Computer Vision and Pattern Recognition}, 2016, pp. 5157--5164.

\bibitem{kang2020structured}
Z.~Kang, C.~Peng, Q.~Cheng, X.~Liu, X.~Peng, Z.~Xu, and L.~Tian, ``Structured
  graph learning for clustering and semi-supervised classification,''
  \emph{Pattern Recognition}, vol. 110, p. 107627, 2021.

\bibitem{xiao2015robust}
S.~Xiao, M.~Tan, D.~Xu, and Z.~Y. Dong, ``Robust kernel low-rank
  representation,'' \emph{IEEE transactions on neural networks and learning
  systems}, vol.~27, no.~11, pp. 2268--2281, 2015.

\bibitem{ma2020towards}
Z.~Ma, Z.~Kang, G.~Luo, L.~Tian, and W.~Chen, ``Towards clustering-friendly
  representations: Subspace clustering via graph filtering,'' in
  \emph{Proceedings of the 28th ACM International Conference on Multimedia},
  2020, pp. 3081--3089.

\bibitem{xie2016unsupervised}
J.~Xie, R.~Girshick, and A.~Farhadi, ``Unsupervised deep embedding for
  clustering analysis,'' in \emph{International conference on machine
  learning}, 2016, pp. 478--487.

\bibitem{guo2017improved}
X.~Guo, L.~Gao, X.~Liu, and J.~Yin, ``Improved deep embedded clustering with
  local structure preservation.'' in \emph{IJCAI}, 2017, pp. 1753--1759.

\bibitem{yang2017towards}
B.~Yang, X.~Fu, N.~D. Sidiropoulos, and M.~Hong, ``Towards k-means-friendly
  spaces: Simultaneous deep learning and clustering,'' in \emph{Proceedings of
  the 34th International Conference on Machine Learning-Volume 70}.\hskip 1em
  plus 0.5em minus 0.4em\relax JMLR. org, 2017, pp. 3861--3870.

\bibitem{zhang2019ae2}
C.~Zhang, Y.~Liu, and H.~Fu, ``Ae2-nets: Autoencoder in autoencoder networks,''
  in \emph{Proceedings of the IEEE Conference on Computer Vision and Pattern
  Recognition}, 2019, pp. 2577--2585.

\bibitem{kang2020structure}
Z.~Kang, X.~Lu, Y.~Lu, c.~Peng, W.~Chen, and Z.~Xu, ``Structure learning with
  similarity preserving,'' \emph{Neural Networks}, vol. 129, pp. 138--148,
  2020.

\bibitem{huang2014deep}
P.~Huang, Y.~Huang, W.~Wang, and L.~Wang, ``Deep embedding network for
  clustering,'' in \emph{2014 22nd International Conference on Pattern
  Recognition}.\hskip 1em plus 0.5em minus 0.4em\relax IEEE, 2014, pp.
  1532--1537.

\bibitem{caron2018deep}
M.~Caron, P.~Bojanowski, A.~Joulin, and M.~Douze, ``Deep clustering for
  unsupervised learning of visual features,'' in \emph{Proceedings of the
  European Conference on Computer Vision (ECCV)}, 2018, pp. 132--149.

\bibitem{jiang2017variational}
Z.~Jiang, Y.~Zheng, H.~Tan, B.~Tang, and H.~Zhou, ``Variational deep embedding:
  an unsupervised and generative approach to clustering,'' in \emph{Proceedings
  of the 26th International Joint Conference on Artificial Intelligence}, 2017,
  pp. 1965--1972.

\bibitem{ghasedi2017deep}
K.~Ghasedi~Dizaji, A.~Herandi, C.~Deng, W.~Cai, and H.~Huang, ``Deep clustering
  via joint convolutional autoencoder embedding and relative entropy
  minimization,'' in \emph{Proceedings of the IEEE international conference on
  computer vision}, 2017, pp. 5736--5745.

\bibitem{jabi2019deep}
M.~Jabi, M.~Pedersoli, A.~Mitiche, and I.~B. Ayed, ``Deep clustering: On the
  link between discriminative models and k-means,'' \emph{IEEE Transactions on
  Pattern Analysis and Machine Intelligence}, 2019.

\bibitem{harchaoui2017deep}
W.~Harchaoui, P.-A. Mattei, and C.~Bouveyron, ``Deep adversarial gaussian
  mixture auto-encoder for clustering,'' in \emph{Workshop of the 5th
  International Conference on Learning Representations (ICLR).}, 2017.

\bibitem{chen2016infogan}
X.~Chen, Y.~Duan, R.~Houthooft, J.~Schulman, I.~Sutskever, and P.~Abbeel,
  ``Infogan: Interpretable representation learning by information maximizing
  generative adversarial nets,'' in \emph{Advances in neural information
  processing systems}, 2016, pp. 2172--2180.

\bibitem{mrabah2019adversarial}
N.~Mrabah, M.~Bouguessa, and R.~Ksantini, ``Adversarial deep embedded
  clustering: on a better trade-off between feature randomness and feature
  drift,'' \emph{arXiv preprint arXiv:1909.11832}, 2019.

\bibitem{shah2017robust}
S.~A. Shah and V.~Koltun, ``Robust continuous clustering,'' \emph{Proceedings
  of the National Academy of Sciences}, vol. 114, no.~37, pp. 9814--9819, 2017.

\bibitem{wang2017feature}
S.~Wang, Z.~Ding, and Y.~Fu, ``Feature selection guided auto-encoder,'' in
  \emph{Thirty-First AAAI Conference on Artificial Intelligence}, 2017.

\bibitem{kang2020relation}
Z.~Kang, X.~Lu, J.~Liang, K.~Bai, and Z.~Xu, ``Relation-guided representation
  learning,'' \emph{Neural Networks}, vol. 131, pp. 93--102, 2020.

\bibitem{mrabah2019deep}
N.~Mrabah, N.~M. Khan, R.~Ksantini, and Z.~Lachiri, ``Deep clustering with a
  dynamic autoencoder: From reconstruction towards centroids construction,''
  \emph{arXiv preprint arXiv:1901.07752}, 2019.

\bibitem{peng2018structured}
X.~Peng, J.~Feng, S.~Xiao, W.-Y. Yau, J.~T. Zhou, and S.~Yang, ``Structured
  autoencoders for subspace clustering,'' \emph{IEEE Transactions on Image
  Processing}, vol.~27, no.~10, pp. 5076--5086, 2018.

\bibitem{li2017projective}
J.~Li and H.~Liu, ``Projective low-rank subspace clustering via learning deep
  encoder,'' in \emph{IJCAI}, 2017.

\bibitem{ji2017deep}
P.~Ji, T.~Zhang, H.~Li, M.~Salzmann, and I.~Reid, ``Deep subspace clustering
  networks,'' in \emph{Advances in Neural Information Processing Systems},
  2017, pp. 24--33.

\bibitem{zhou2018deep}
P.~Zhou, Y.~Hou, and J.~Feng, ``Deep adversarial subspace clustering,'' in
  \emph{Proceedings of the IEEE Conference on Computer Vision and Pattern
  Recognition}, 2018, pp. 1596--1604.

\bibitem{kheirandishfard2020multi}
M.~Kheirandishfard, F.~Zohrizadeh, and F.~Kamangar, ``Multi-level
  representation learning for deep subspace clustering,'' in \emph{The IEEE
  Winter Conference on Applications of Computer Vision}, 2020, pp. 2039--2048.

\bibitem{abavisani2018deep}
M.~Abavisani and V.~M. Patel, ``Deep multimodal subspace clustering networks,''
  \emph{IEEE Journal of Selected Topics in Signal Processing}, vol.~12, no.~6,
  pp. 1601--1614, 2018.

\bibitem{zhu2019multi}
P.~Zhu, B.~Hui, C.~Zhang, D.~Du, L.~Wen, and Q.~Hu, ``Multi-view deep subspace
  clustering networks,'' \emph{arXiv preprint arXiv:1908.01978}, 2019.

\bibitem{lee2013pseudo}
D.-H. Lee, ``Pseudo-label: The simple and efficient semi-supervised learning
  method for deep neural networks,'' in \emph{Workshop on challenges in
  representation learning, ICML}, vol.~3, 2013, p.~2.

\bibitem{chang2017deep}
J.~Chang, L.~Wang, G.~Meng, S.~Xiang, and C.~Pan, ``Deep adaptive image
  clustering,'' in \emph{Proceedings of the IEEE international conference on
  computer vision}, 2017, pp. 5879--5887.

\bibitem{yang2016joint}
J.~Yang, D.~Parikh, and D.~Batra, ``Joint unsupervised learning of deep
  representations and image clusters,'' in \emph{Proceedings of the IEEE
  Conference on Computer Vision and Pattern Recognition}, 2016, pp. 5147--5156.

\bibitem{bengio2007greedy}
Y.~Bengio, P.~Lamblin, D.~Popovici, and H.~Larochelle, ``Greedy layer-wise
  training of deep networks,'' in \emph{Advances in neural information
  processing systems}, 2007, pp. 153--160.

\bibitem{chen2017unsupervised}
D.~Chen, J.~Lv, and Y.~Zhang, ``Unsupervised multi-manifold clustering by
  learning deep representation,'' in \emph{Workshops at the Thirty-First AAAI
  Conference on Artificial Intelligence}, 2017.

\bibitem{jiang2018learn}
Y.~Jiang, Z.~Yang, Q.~Xu, X.~Cao, and Q.~Huang, ``When to learn what: Deep
  cognitive subspace clustering,'' in \emph{Proceedings of the 26th ACM
  international conference on Multimedia}, 2018, pp. 718--726.

\bibitem{jing2019self}
L.~Jing and Y.~Tian, ``Self-supervised visual feature learning with deep neural
  networks: A survey,'' \emph{arXiv preprint arXiv:1902.06162}, 2019.

\bibitem{kolesnikov2019revisiting}
A.~Kolesnikov, X.~Zhai, and L.~Beyer, ``Revisiting self-supervised visual
  representation learning,'' in \emph{Proceedings of the IEEE conference on
  Computer Vision and Pattern Recognition}, 2019, pp. 1920--1929.

\bibitem{kilinc2018learning}
O.~Kilinc and I.~Uysal, ``Learning latent representations in neural networks
  for clustering through pseudo supervision and graph-based activity
  regularization,'' in \emph{ICLR}, 2018.

\bibitem{oord2018representation}
A.~v.~d. Oord, Y.~Li, and O.~Vinyals, ``Representation learning with
  contrastive predictive coding,'' \emph{arXiv preprint arXiv:1807.03748},
  2018.

\bibitem{wu2019deep}
J.~Wu, K.~Long, F.~Wang, C.~Qian, C.~Li, Z.~Lin, and H.~Zha, ``Deep
  comprehensive correlation mining for image clustering,'' in \emph{The IEEE
  International Conference on Computer Vision (ICCV)}, October 2019.

\bibitem{zhang2019self}
J.~Zhang, C.-G. Li, C.~You, X.~Qi, H.~Zhang, J.~Guo, and Z.~Lin,
  ``Self-supervised convolutional subspace clustering network,'' in
  \emph{Proceedings of the International Conference on Computer Vision}, 2019,
  pp. 5473--5482.

\bibitem{kang2020robust}
Z.~{Kang}, H.~{Pan}, S.~C.~H. {Hoi}, and Z.~{Xu}, ``Robust graph learning from
  noisy data,'' \emph{IEEE Transactions on Cybernetics}, vol.~50, no.~5, pp.
  1833--1843, 2020.

\bibitem{bromley1994signature}
J.~Bromley, I.~Guyon, Y.~LeCun, E.~S{\"a}ckinger, and R.~Shah, ``Signature
  verification using a" siamese" time delay neural network,'' in \emph{Advances
  in neural information processing systems}, 1994, pp. 737--744.

\bibitem{luo2018smooth}
Y.~Luo, J.~Zhu, M.~Li, Y.~Ren, and B.~Zhang, ``Smooth neighbors on teacher
  graphs for semi-supervised learning,'' in \emph{Proceedings of the IEEE
  conference on computer vision and pattern recognition}, 2018, pp. 8896--8905.

\bibitem{liu2013robust}
G.~Liu, Z.~Lin, S.~Yan, J.~Sun, Y.~Yu, and Y.~Ma, ``Robust recovery of subspace
  structures by low-rank representation,'' \emph{IEEE transactions on pattern
  analysis and machine intelligence}, vol.~35, no.~1, pp. 171--184, 2013.

\bibitem{patel2014kernel}
V.~M. Patel and R.~Vidal, ``Kernel sparse subspace clustering,'' in \emph{Image
  Processing (ICIP), 2014 IEEE International Conference on}.\hskip 1em plus
  0.5em minus 0.4em\relax IEEE, 2014, pp. 2849--2853.

\bibitem{you2016scalable}
C.~You, D.~Robinson, and R.~Vidal, ``Scalable sparse subspace clustering by
  orthogonal matching pursuit,'' in \emph{Proceedings of the IEEE conference on
  computer vision and pattern recognition}, 2016, pp. 3918--3927.

\bibitem{ji2014efficient}
P.~Ji, M.~Salzmann, and H.~Li, ``Efficient dense subspace clustering,'' in
  \emph{Applications of Computer Vision (WACV), 2014 IEEE Winter Conference
  on}.\hskip 1em plus 0.5em minus 0.4em\relax IEEE, 2014, pp. 461--468.

\bibitem{fard2018deep}
M.~M. Fard, T.~Thonet, and E.~Gaussier, ``Deep $ k $-means: Jointly clustering
  with $ k $-means and learning representations,'' \emph{arXiv preprint
  arXiv:1806.10069}, 2018.

\bibitem{yang2019deep}
X.~Yang, C.~Deng, F.~Zheng, J.~Yan, and W.~Liu, ``Deep spectral clustering
  using dual autoencoder network,'' in \emph{Proceedings of the IEEE Conference
  on Computer Vision and Pattern Recognition}, 2019, pp. 4066--4075.

\bibitem{kingma2014adam}
D.~P. Kingma and J.~L. Ba, ``Adam: Amethod for stochastic optimization,'' in
  \emph{Proc. 3rd Int. Conf. Learn. Representations}, 2014.

\bibitem{peng2020deep}
X.~Peng, J.~Feng, J.~T. Zhou, Y.~Lei, and S.~Yan, ``Deep subspace clustering,''
  \emph{IEEE transactions on neural networks and learning systems}, vol.~31,
  no.~12, pp. 5509--5521, 2020.

\bibitem{zhang2018scalable}
T.~Zhang, P.~Ji, M.~Harandi, R.~Hartley, and I.~Reid, ``Scalable deep
  k-subspace clustering,'' in \emph{Asian Conference on Computer Vision}.\hskip
  1em plus 0.5em minus 0.4em\relax Springer, 2018, pp. 466--481.

\bibitem{maaten2008visualizing}
L.~v.~d. Maaten and G.~Hinton, ``Visualizing data using t-sne,'' \emph{Journal
  of machine learning research}, vol.~9, no. Nov, pp. 2579--2605, 2008.

\end{thebibliography}

\end{document}